\documentclass[sigconf]{acmart}

\usepackage[ruled,vlined,noend]{algorithm2e}
\usepackage{algorithmic}
\usepackage{multirow}

\usepackage{todonotes}

\usepackage{subfig}

\AtBeginDocument{%
  \providecommand\BibTeX{{%
    \normalfont B\kern-0.5em{\scshape i\kern-0.25em b}\kern-0.8em\TeX}}}

\copyrightyear{2026}
\acmYear{2026}
\setcopyright{cc}
\setcctype{by-nc-nd}

\begin{document}


\title{A Seed for Privacy - semi-automatic privacy-revealing data detection in databases and data streams}

\author{He Gu}
\author{Thomas Plagemann}
\author{Vera Goebel}
\email{heg, plageman, goebel@ifi.uio.no}
\affiliation{%
  \institution{University of Olso}
  \streetaddress{Boks 1072 Blindern}
  \city{Oslo}
  \country{Norway}
  \postcode{0316}
}


\begin{abstract}
Sharing databases and data streams imposes the danger of revealing private information in the form of complex events which can comprise individual data elements and their combinations. Identifying these privacy-revealing complex events is crucial for preserving privacy while maintaining data utility. However, data producers often lack the expertise to comprehensively identify these events, which undermines many state-of-the-art privacy-preserving mechanisms that rely on accurate event labeling. To address this challenge, we developed pArborist - a tool that can semi-automatically create a set of queries to identify and label privacy-revealing complex events in both static datasets and dynamic data streams, guided by the privacy requirements of the data producer. pArborist uses the schema of the database or data stream combined with initial input from the data producer, i.e., seed queries. From each seed query, pArborist grows a tree containing all possible syntactically correct queries, constrained by an upper limit on computational resources. Following this growing phase, the tree is refined by eliminating queries that lack correlation to the seed or are conditionally independent of the seed. Our evaluation indicates that pArborist achieves overall recall of $90\%$ and precision of $93\%$ in finding privacy-revealing queries, and this significantly surpasses the state-of-the-art approach FQID. In data stream processing experiments, pArborist introduces a delay of approximately 1.3 ms following an average warm-up period of 920 ms. The experiments also show that pArborist can automatically detect privacy-revealing complex events according to GDPR.
\end{abstract}





\maketitle
\section{Introduction}\label{intro}
Data sharing is an essential procedure for most data-driven applications and research, which raises great concerns about potential privacy disclosures. A typical approach to protecting private information is anonymization. In practice, anonymization is achieved by protecting identifiers and quasi-identifier. Identifiers denote the attributes that reveal the identity of an individual, while quasi-identifiers are the combinations of certain attributes that jointly reveal the identity of an individual. Detecting identifiers and quasi-identifiers during anonymization is a nontrivial research problem and is still a subject to research \cite{podlesny2023quasi, hua2021method, carvalho2022survey}.

However, anonymization is not always suitable for privacy protections. Consider the case of elderly people living alone. Health monitoring devices continuously send information about the daily activities of the elderly to health personnel to determine whether they need assistance. In such cases, the devices and their users have a strong incentive to share the data without concealing the identity of users. However, there may be some particular activities or sequences of activities that the elderly do not want to share, as they are not related to health monitoring but may reveal certain private information. Therefore, there is a great demand in protecting general private information rather than solely concealing the identifiers and quasi-identifiers. There exist multiple studies for general privacy protection. Some approaches equally protect all shared data including non-privacy-revealing data, usually by introducing additional noise to the original data values. This imposes a significant decrease on data utility for data-driven applications, e.g., health monitoring. Some other studies have proposed improved approaches that apply customized privacy protection focusing on privacy-revealing data and reduce the damage on non-privacy-revealing data. For example, Palanisamy et al. \cite{palanisamy2018preserving} leverage `the power of events' \cite{luckham2002power} and use the concept of complex events to focus on the obfuscation process of a privacy-preserving mechanism (PPM) on those complex events that reveal private information (privacy-revealing complex events).

\begin{figure}
\centering
\includegraphics[width=0.47\textwidth]{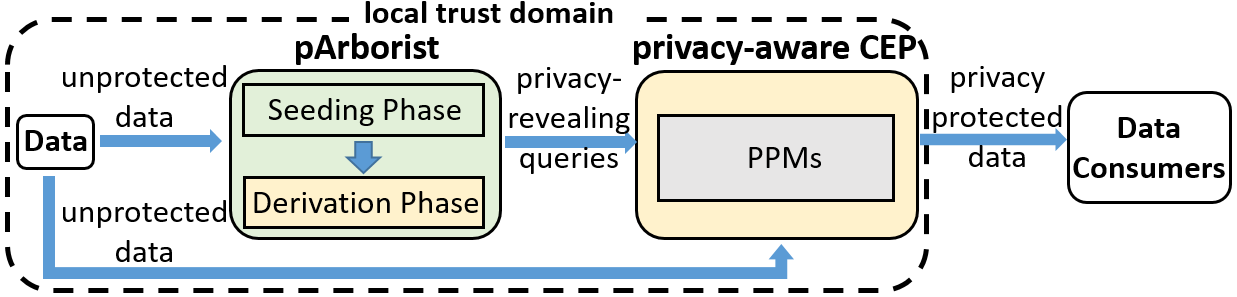}%
\vspace{-7pt}
\caption{The use case of pArborist supporting the application of state-of-the-art PPMs for privacy-aware CEP systems.}
\label{pandothers}
\end{figure}

Having strong privacy protection and high data utility at the same time is an important improvement. All these improved PPMs assume that the privacy-revealing data is already labeled, so that customized privacy protection can be applied. However, in practice, this is usually not guaranteed, and the detection of privacy-revealing data is yet an unsolved challenge. Although there exist generally accepted regulations to define privacy-revealing data, e.g., the General Data Protection Regulation (GDPR) \cite{regulation2018general}, the corresponding implementations to apply these regulations on specific databases or data streams are still in their infancy. In addition, specific data producers may have specific privacy requirements, which can overturn the general regulations. Considering that data producers may not possess sufficient privacy expertise, there is a great danger in privacy disclosures due to imprecisely labeled privacy-revealing data given by data producers. In conclusion, to enable enhanced PPMs and empower subsequent data sharing, there are three unsolved challenges considering the detection of privacy-revealing data: (1) how to implement generally accepted privacy-preserving regulations, e.g., GDPR, on databases and data streams, (2) how to precisely define the privacy requirements of data producers, and (3) how to identify the corresponding privacy-revealing data according to the privacy requirements.

This paper presents the \textbf{pArborist} (privacy tree-specialist) approach that semi-automatically creates queries to identify and label privacy-revealing complex events in given databases or data streams, including (1) identifiers and quasi-identifiers, and (2) complex events that data producers regard as private. pArborist is specifically designed to support data producers who possess limited privacy expertise. By default, pArborist uses legal regulations, such as GDPR, if no privacy requirements are provided. To achieve more personalized solutions, data producers need to provide a few \textbf{seed queries} (seeds) to pArborist as instructions, which are one or more queries that reveal private information. They can, for example, reveal the locations for particular confidential GPS coordinates. pArborist develops trees of syntactically correct queries and uses seed queries to detect and verify new queries that reveal the same private information as the seed queries, e.g., receipts in hotels that reveal the presence at the same confidential GPS coordinates. Since new information is continuously generated in real-time data streams and privacy-revealing complex events can be arbitrarily complex, it is infeasible to enumerate all such events in advance. Therefore, pArborist converges towards finding a comprehensive set of privacy-revealing queries. Figure \ref{pandothers} illustrates how these queries can be used to achieve a privacy-aware Complex Event Processing (CEP) system with state-of-the-art PPMs that require labeled privacy-revealing complex events, e.g., \cite{gu2023differential, palanisamy2018preserving, lotfian2025pattern}, for given databases or data streams and specific privacy requirements. This paper makes the following contributions.
\begin{itemize}
    \item We propose a novel semi-automatic approach pArborist to identify and label privacy-revealing complex events. It takes as inputs either seed queries specified by data producers or legal regulations and produces a set of derived privacy-revealing queries that can retrieve privacy-revealing complex events in given databases or data streams.
    \item We design experiments to evaluate recall, precision, efficiency, and practicality of privacy detection approaches, based on three real-world datasets. We have demonstrated the performance of pArborist based on the state-of-the-art approach FQID \cite{podlesny2023quasi}. We present an application of pArborist that automatically identifies potential violations of GDPR in a given database or data stream.
    \item We label public datasets with privacy-revealing complex events and publish the labeled data with an implementation of pArborist.
\end{itemize}

The rest of the paper is organized as follows: We introduce the related works in Section 2 and the problem statement in Section 3. We describe pArborist in Section 4, and in Section 5, we detail the derivation algorithms in pArborist. We evaluate pArborist in Section 6. Section 7 concludes this paper.


\section{Related works} \label{related works}
The need to support individuals or organizations to identify privacy-revealing data when sharing or publishing data has been recognized. Some studies propose solutions based on security measures \cite{shu2015privacy, nan2015uipicker}. They may deploy a centralized monitoring system over both data producers and consumers, so that the unknown privacy-revealing data can be labeled once a privacy disclosure is detected \cite{nan2015uipicker}. Furthermore, some studies \cite{shu2015privacy} can detect unknown privacy-revealing data by inserting identifiable marks into known privacy-revealing data and tracking these marks until there are privacy disclosures. The solutions based on security measures usually aim to discover new privacy-revealing data after they have been leaked, which is therefore less preferred for privacy protection scenarios. 

In addition to security measures, some techniques can identify unknown privacy-revealing data prior to disclosures \cite{caliskan2014privacy, tesfay2019privacybot}. Machine learning models \cite{kopeykina2019automatic, caliskan2014privacy, mehdy2019privacy, tesfay2019privacybot}, e.g., large language models, have been applied and proven effective in detecting privacy in natural language scenarios \cite{caliskan2014privacy, mehdy2019privacy, tesfay2019privacybot}, e.g., discovering privacy risks in posted tweets \cite{caliskan2014privacy}. However, these models are highly limited to natural language scenarios, and there are concerns about their interpretability and energy consumption if they are required to independently detect privacy-revealing data for each data producer and each use case. Their generalization raises concerns for users because of the lack of training data, e.g., datasets with well-labeled privacy-revealing data. Another typical solution is to construct a dictionary for already specified privacy-revealing data \cite{vasalou2011privacy}, so that the unspecified privacy-revealing data can be discovered by looking up the dictionary. However, its generalization presents a challenge to both human resources and computing resources, since most elements of a privacy dictionary are manually selected, data-driven, and can hardly be applied to distinct use cases. Furthermore, many aspects of privacy are highly dependent on the preference of each individual data producer, leading to a great number of personalized dictionaries and their continuous updates for new use cases.

Multiple studies have considered a more general approach than machine learning models and privacy dictionary solutions. They aim to detect unspecified sensitive attributes in any given database by applying statistical analytics. However, most relevant studies focus only on a rather narrower privacy category, e.g., identifiers and quasi-identifiers for anonymization \cite{podlesny2023quasi, hua2021method, carvalho2022survey}. The challenges in such cases are mostly related to the optimization of computing resource consumption, instead of the detection techniques for unspecified privacy-revealing data. These solutions are limited to anonymization and can hardly be applied for other privacy categories than identifiers and quasi-identifiers. 

We take into account the interpretability, generalization, efficiency, and resource consumption problems in the related works and propose pArborist. pArborist (1) is agnostic to privacy categories, (2) consumes an affordable amount of computing resources which enables real-time data stream processing, and (3) requires few human resources and little privacy expertise by almost automatically detecting privacy-revealing data for data producers. 

\section{Problem statement} \label{PS}
Anonymization is a generally accepted approach to protect private information in shared data by masking the identities of individuals, which, however, may be inadequate in many cases as explained in Section \ref{intro}. Consequently, this work considers privacy-revealing data to comprise not only these identities of individuals, but also any information that an individual does not want to share. 

Privacy-revealing data are usually regulated by laws, e.g., GDPR, while data producers can additionally label data as either privacy-revealing or not. For example, some data producers may regard their health status as private, while others share them on social media. In a given database or data stream, privacy-revealing data can be practically represented as privacy-revealing complex events, formed by multiple basic events that together expose higher-level private information. Advanced PPMs protect privacy while maintaining high data utility if they get data with labeled privacy-revealing complex events. Data producers have the authority to label their data, but their lack of privacy expertise can lead to inaccurate labels. For example, although data producers label their hotel addresses as privacy-revealing, they may not realize that their Internet connections in hotels can also reveal hotel addresses (Hotel Internet Example\footnote{We distinguish clearly between data (e.g., the hotel address or Internet connection) and information those data reveal (e.g., the location).}). This inaccurate labeling may have critical consequences, as it falsifies the fundamental assumption of multiple state-of-the-art PPMs \cite{palanisamy2018preserving, gu2023differential} that all privacy-revealing complex events have been accurately labeled, and this can greatly downgrade the performance of these PPMs w.r.t. data utilities and privacy protections. 

Therefore, it is crucial to assist data producers in labeling privacy-revealing complex events based on their privacy requirements, which is non-trivial according to existing studies \cite{podlesny2023quasi} and is the research problem of this work. This problem consists of two sub-problems: (1) to support data producers to formulate their privacy requirements, and (2) to label a comprehensive set of privacy-revealing complex events according to the formulated requirements. 

Sub-problem (1) involves two scenarios: (1) data producers do not have specific privacy requirements, or (2) data producers have privacy requirements, but they may unintentionally omit certain requirements. For example, although hotel addresses are labeled as privacy-revealing, data producers may unintentionally omit to label hotel restaurants as privacy-revealing as well. Sub-problem (2) leads to two main challenges: (1) it is non-trivial to label privacy-revealing complex events that are not explicitly specified by privacy requirements but are correlated to these requirements, and (2) the search space for potential privacy-revealing complex events is infinite because complex events can be arbitrarily complex and new complex events are continuously generated in real-time data streams.

\section{pArborist Approach}
It is unrealistic to enumerate all privacy-revealing complex events in advance for real-time data streams, and it is inefficient to label each specific complex event in databases. Therefore, pArborist constructs privacy-revealing queries instead of labeling each complex event. A privacy-revealing query defines a universal condition which can retrieve all corresponding privacy-revealing complex events. For example, regarding a privacy-revealing query ``when Bob is in a Hilton hotel'', a corresponding complex event can be either ``when Bob is in a Hilton hotel in Japan'', ``in the UK'', or ``in Denmark''. 

Given a database or a data stream, the privacy-revealing queries constructed by pArborist are derived from seed queries that originate from the privacy requirements of data producers. For convenience, this paper regards both databases and data streams as \textbf{sets of data} $S_d$, which can be finite or countably infinite  based on the set theory. The pArborist approach is performed in two phases: (1) In the \textbf{seeding phase}, we support data producers to formulate their privacy requirements into seed queries, i.e., seeds. (2) In the \textbf{derivation phase}, we collect for each seed a comprehensive set of potentially relevant queries and identify those that reveal the same private information as the seed. The resulting privacy-revealing queries can subsequently be directly used to retrieve and label privacy-revealing complex events in databases or data streams. 

In this section, we present the seeding phase and give an overview of the derivation phase, including its core principles and main steps, while its detailed algorithms are presented in Section \ref{DPA}.

\subsection{Seeding Phase} \label{FofS}
The goal of the seeding phase is to assist data producers in formulating seed queries, even if they lack extensive privacy expertise. This is performed in three stages:
\begin{enumerate}
    \item Formulate the privacy requirement of a data producer into an initial seed query.\footnote{If the data producer formulates more than one initial seed query, the seeding phase is individually performed for each of them.}
    \item From the initial seed, infer other potential seeds that may reveal the same private information, but are unintentionally omitted by the data producer.
    \item Present the inferred seed queries to the data producer for confirmation.
\end{enumerate}
All confirmed queries, including the initial seed query of the data producer, are considered seed queries and are the input for the derivation phase. The details of the seeding phase are described below in five steps.

In Step 1, we collect a comprehensive list of attributes and predicates of queries based on the schema of the given set of data and the applied query language. We combine each attribute and one of its syntactically correct predicates into an \textbf{Attribute-Function (A-F) pair} $(A, f_{O^A_i}(\mathbf{x}_i))$, consisting of a selected attribute and a function $f_{O^A_i}$ that denotes the operator $O^A_i$ applied to $A$ with parameters $\mathbf{x} = (x_1, x_2, ..., x_n)$. For example, a predicate that selects data where the customer's name equals Bob is combined with its attribute into an A-F pair $(\mathit{name}, f_{\mathit{eq}}(\mathit{Bob}))$. In addition to basic predicates, clauses  and aggregations can also be expressed by A-F pairs with complex attributes or functions. For example, Join clause between tables $t_1$ and $t_2$ on attribute $\mathit{t}_1.A = \mathit{t}_2.B$ can be converted to $(\mathit{ALL}, f_{\mathit{eqjoin}}(A_{\mathit{t}_1}, B_{\mathit{t}_2})$. GROUP BY clause on attribute $A$ with SUM function on attribute $B$ can be converted into $(B, f_{\mathit{sum}}(f_{\mathit{groupBy}}(A)))$. This paper considers such a compound A-F pair as $n$ A-F pairs if the compound pair involves $n$ attributes, because the depth of trees built by pArborist is determined by the number of attributes involved in a query. We use A-F pairs instead of directly using predicates because (1) A-F pairs support user-defined operators and functions that take multiple parameters as input, (2) A-F pairs can be naturally indexed by attributes, which simplifies iteration and search in our algorithms, (3) A-F pairs simplify the presentation of algorithms and their implementation, and (4) A-F pairs can be mapped to various query languages and data management systems.

In Step 2, the data producer formulates privacy requirements into privacy-revealing queries, i.e., seed queries, based on the list of A-F pairs. Consider a Check-in example where the check-in data of Bob contains his private information, i.e., he went to hotel H on April 10th. Its corresponding query, 
\begin{equation*}
\begin{aligned}
    &\mathit{SELECT}\;* \\
    &\mathit{FROM}\;\text{Check-in} \\
    &\mathit{WHERE}\;\text{name} = \text{`Bob'}\;\mathit{AND}\;\text{venueName} = \text{`H'}\;\mathit{AND}\; \\
    &\quad \quad \quad \quad \text{date} <= 04102359\;\mathit{AND}\;\text{date} >= 04100000,
\end{aligned}
\end{equation*}
can be constructed based on a set of three A-F pairs, i.e., $\{(\mathit{name}, \\f_{\mathit{eq}}(\mathit{Bob}))$, $(\mathit{venue}, f_{\mathit{eq}}(H))$, $(\mathit{date}, f_{\mathit{within}}(04100000, 04102359))\}$. The design of A-F pairs enables data producers with little database expertise to construct their queries, since an A-F pair can be automatically translated and presented to the data producer as an expression in natural language.

In the next steps, we aim to infer from the data producer's initial seed query other possible seed queries that may reveal the same private information, but are unintentionally omitted by the data producer. In Step 3, we populate a dictionary $D$ consisting of A-F pairs that have the same attribute, function, number of parameters, and the values of at least half of their parameters as each pair in the seed set. For example, the A-F pair $(\mathit{date}, f_{\mathit{within}}(04100000, 04102359))$ in the seed set shares with $(\mathit{date}, f_{\mathit{within}}(04102359, 04112359))$ one of their two parameters $04102359$ of the same function $f_{\mathit{within}}$.

In Step 4, we generate candidate seed queries that might be unintentionally omitted by the data producer. For a seed set, we replace one of its A-F pair by another possibly relevant pair in the dictionary $D$ to form a new seed set. We repeat this procedure for each pair in $D$ to generate multiple candidate seed sets.

In Step 5, we present all generated seed sets to the data producer for confirmation. If there are exceedingly many candidates, we merge queries that differ only in the value of one predicate by replacing the value with a wildcard, in which we ask the data producer to specify a value. The number of candidates that trigger the wildcards is configurable. We propose six based on our experience in the experiments. In the context of the Check-in example, the private information is ``Bob went to hotel H on April 10th''. To construct candidate seed queries, we first replace the name Bob with other possibly relevant names. If there exist two names, e.g., David and James, we replace Bob with David and James to form two new queries for confirmation. If six or more names appear, we replace Bob with a wildcard and ask for a manual input name from the data producer to continue the construction.

In practice, we consider two critical challenges. First, the data producer may have little knowledge about the construction of queries. In such cases, we translate all A-F pairs into expressions in natural language and present them to the data producer for the confirmation of seed queries. Second, a data producer might unintentionally omit a whole category of privacy-revealing complex events, e.g., all location-based data. Since the inference of seed queries depends on the initial seed given by data producers, the seeding phase cannot infer any seed query in an omitted privacy category. To address such cases, we maintain a list of default privacy categories with their suggested seed queries shared among all users of pArborist. If the data producer does not provide any seed query targeting a particular privacy category, we notify the data producer and verify whether the omission was intentional. If not, we ask for additional input from data producers and present to them the suggested seed queries in the category as a reference.

\subsection{Derivation Phase}\label{FofD}
In the derivation phase, we derive other privacy-revealing queries from the seed queries based on a tree structure in two stages:
\begin{enumerate}
    \item \textbf{Tree Growth}: We generate a comprehensive set of syntactically valid queries for the given set of data by growing a tree that explores possible query formulations \footnote{The term ``seed'' might imply that a seed is necessary for Tree Growth, which is not true. The seeds are only necessary for Node Bypassing.}. The grown tree at this stage can be shared among data producers.
    \item \textbf{Node Bypassing}: We refer to all queries in the tree that are correlated and conditionally dependent with the seed query as valid queries (valid nodes). We reconstruct the tree with only valid nodes while removing all invalid nodes. This reconstructed tree can only be used by this data producer.
\end{enumerate}
The nodes from all trees combined are the derived privacy-revealing queries. When we refer to the fact that two queries are correlated or conditionally dependent, we indicate that the answers of these queries are correlated or conditionally dependent for a set of data. The implementation is performed in four steps (see Algorithm \ref{derivation phase}). 
\begin{algorithm}
\SetAlgoLined
 1. Reuse the list of A-F pairs used in the seeding phase.\\
 \For{each seed query constructed in the seeding phase}{
    2. \textbf{Tree Growth:}\\
    \quad 2.1. Decompose the seed into a set of $n$ A-F pairs. \\
    \uIf{there is no default tree}{
        2.2.1. Use \textbf{Growing Method} to grow a default tree; \\
    }
    \uElseIf{the seed is included in the default tree}{
        2.2.2. Make a copy of the default tree for the seed;\\
    }
    \Else{
        2.2.3. Apply the \textbf{Striking Method} to grow a tree; \\
    }
    3. \textbf{Node Bypassing:}\\
    \For{each node in the tree}{
        \uIf{the node is not correlated with the seed}{
            3.1. Remove the node and connect its parent node to its child nodes;\\
        }
    }
    \For{each node in the new tree}{
        \uIf{the node is conditionally independent with the seed}{
        3.2. Remove the node and connect its parent node to its child nodes;
        }
    }
}
4. Collect all nodes from all trees.\\
 \caption{Derivation Phase.}
 \label{derivation phase}
\end{algorithm}

In Step 1, we reuse the list of all syntactically correct A-F pairs collected in the seeding phase as the building blocks for queries. In Step 2, we generate a comprehensive set of syntactically valid queries by growing a tree. By default, we apply the growing method to grow the tree with the given amount of computational resource. If we assume that the grown tree is of depth $d$, all seed queries consisting of no more than $d$ A-F pairs can directly utilize a copy of this tree in the next steps, without repeatedly executing the growing method. Meanwhile, for a seed query consisting of more than $d$ A-F pairs, the tree constructed by the growing method cannot be used, and we utilize the striking method. In Step 3, we analyze all nodes in the constructed tree and verify those that can reveal the same private information as the seed query, based on correlation and conditional dependence. We bypass the nodes that have not been verified. In Step 4, we collect nodes from all trees and deliver them to data producers when they require.

Note that although Tree Growth is a brute-force procedure based on combinatorics, it is performed only once for a given data schema. Additionally, this grown tree, which has not been processed at Node Bypassing stage, can be shared among all data producers whose data adhere to the same schema because it does not contain any private information. The generalization of a tree can be performed by a third party. For example, in a crowdsensing campaign, the organizer could generate the tree and distribute it to all data producers. Data producers would only need to perform the Node Bypassing step on their own trusted devices, which requires low computing power. This is discussed and confirmed by the experiments in Section \ref{evaluation}.
 
\section{Derivation Algorithms} \label{DPA}
In this section, we present the detailed algorithms of the derivation phase for any given seed query, including Tree Growth (Algorithm \ref{derivation phase}, Step 2) and Node Bypassing (Algorithm \ref{derivation phase}, Step 3).

We denote a tree as a countably infinite ordered set $T:=\{t_0, t_1, ..., \\t_n, ...\}$. Its element $t_i$ is the set of all nodes at depth $i$, i.e., $t_i:=\{q^i_1, q^i_2, ..., q^i_{n_i}\}$. Here, $q^i_j$ is a node, i.e., a query represented in the form of A-F pairs, and $n_i$ is the number of nodes at depth $i$. The root of the tree is an empty node, i.e., $t_0=\{q^0_1\}$ and $q^0_1 = \varnothing$. All nodes at the same depth in a tree have the same number of A-F pairs, and their child nodes contain exactly one more A-F pair. We specifically denote the seed node of the tree as $q^s$ for convenience.

\subsection{Tree Growth Algorithms} \label{TGA}
In Tree Growth, we aim to generate a set of all syntactically valid queries by growing a tree $T:=\{t_0, t_1, ..., t_d\}$ with the given computational resource. We propose two methods for growing trees based on combinatorics, i.e. (1) growing method and (2) striking method. By default, we utilize the growing method, which grows the tree from the root node. The striking method is applied if the given computational resource is not sufficient for the growing method to grow a tree that includes the seed query. This method grows a tree from the seed query.

\subsubsection{Growing Method}
The growing method grows a tree starting with an empty root. In an iterative process, new child nodes are generated at the next depth of the tree until the maximum amount of computational resources is consumed, as demonstrated in Algorithm \ref{the growing algorithm}. The tree is initialized as $T=\{t_0\}$, where $t_0=\{q^0_1\}$ and $q^0_1 = \varnothing$. For a node consisting of $i$ A-F pairs, we generate one of its child nodes at depth $i+1$ by adding a new A-F pair to it. By iterating this procedure for all existing nodes until the maximum amount of computational resources is consumed, we construct a tree of its maximum depth $d_{max}$.
\begin{algorithm}
\SetAlgoLined
 1. $L^{\mathit{all}} \leftarrow$ all syntactically correct A-F pairs $\{(A_1, f_{O^{A_1}}(\mathbf{x}_i)), (A_2, f_{O^{A_2}}(\mathbf{x}_i))...,(A_n, f_{O^{A_n}}(\mathbf{x}_i))\}$.\\
 2. $q^0_1 \leftarrow \mathit{node}.\mathit{initialize}(\{\})$; $t_0 \leftarrow \{q^0_1\}$; $T \leftarrow \{t_0\}$.\\
 3. $d_{\mathit{max}} \leftarrow$ estimated maximum depth.\\
 \For{$i \leftarrow 1$; $i \leftarrow i + 1$; $i \leq d_{\mathit{max}}$}{
    4.1. $t_i \leftarrow \{\}$.\\
    \For{$j \leftarrow 0$; $j \leftarrow j + 1$; $j \leq t_{i-1}.size()$}{
        \For{$k \leftarrow 0$; $k \leftarrow k + 1$; $k \leq n$}{
        4.2.1. $q^{\mathit{temp}} \leftarrow$ $q^{i-1}_j$.\\
        4.2.2. $t_i.\mathit{append}(q^{\mathit{temp}}.\mathit{append}((A_k,f_{O^{A_k}}(\mathbf{x}))))$.\\
        }
    }
    4.3. $T.\mathit{append}(t_i)$.\\
 }
 5. Output the grown tree $T$.\\
 \caption{Growing Method.}
 \label{the growing algorithm}
\end{algorithm}
It is beneficial for pArborist to ensure that the most similar queries of a given seed query are included in the tree for further verifications. Consider a seed query consisting of $n$ A-F pairs, its most similar queries usually consist of close to $n$ A-F pairs. Therefore, the tree grown by the growing method is used for all seed queries consisting of less than $d_{\mathit{max}}$ A-F pairs. In such cases, the most similar queries of each seed are included in the tree and will be investigated in the Node Bypassing step to verify whether they reveal the same private information as the seed.

\subsubsection{Striking Method}
For a seed query consisting of more than $d_{\mathit{max}}$ A-F pairs, the tree does not include the most similar queries of the seed. In such cases, the striking method is utilized. As described in Algorithm \ref{the striking algorithm}, this method grows a tree bidirectionally from the given seed query of $n$ A-F pairs at depth $n$. By replacing one of its A-F pairs into another A-F pair, we generate a new node at depth $n$. After generating all syntactically valid nodes at depth $n$ (Step 3), by adding or removing one A-F pair for each of these nodes, we generate new nodes at depth $n+1$ (Steps 4.1 - 4.3) or depth $n-1$ (Steps 4.4.1 - 4.4.4). We iterate this procedure within the given computational resources to construct a tree from depth $n-d$ to depth $n+d$, where $n-d \geq 0$.

The growth of the tree from depth $i$ to depth $i+1$ automatically constructs connections between the child nodes at depth $i+1$ and their parent nodes at depth $i$. However, generating nodes at depth $i-1$ from depth $i$ requires additional operations. For a given node at depth $i$, we iteratively investigate each node at depth $i-1$, and the first node that shares $i-1$ A-F pairs with the given node is its parent nodes (Steps 4.4.4 - 4.5).

\begin{algorithm}
\SetAlgoLined
 1. $L^{all} \leftarrow$ all syntactically correct A-F pairs $\{(A_1, f_{O^{A_1}}(\mathbf{x}_i)), $\\$(A_2, f_{O^{A_2}}(\mathbf{x}_i))...,(A_n, f_{O^{A_n}}(\mathbf{x}_i))\}$; $m \leftarrow q^s.\mathit{size}()$.\\
 2. $t_m \leftarrow \mathit{combGenerator}(\mathit{elements} = L^{\mathit{all}}, size = m)$; \\
 3. $T \leftarrow \{t_m\}$; $d_{\mathit{max}} \leftarrow$ estimated maximum depth.\\
 \For{$i \leftarrow 1$; $i \leftarrow i + 1$; $i \leq d_{\mathit{max}}$}{
    4.1. $d \leftarrow m + i.$; $t_d \leftarrow \{\}$.\\
    \For{$j \leftarrow 1$; $j \leftarrow j + 1$; $j \leq t_{d-1}.\mathit{size}()$}{
        \For{$k \leftarrow 1$; $k \leftarrow k + 1$; $k \leq n$}{
        4.2.1. $q^{\mathit{temp}} \leftarrow$ $q^{d-1}_j$.\\
        4.2.2. $t_d.\mathit{append}(q^{\mathit{temp}}.\mathit{append}((A_k,f_{O^{A_k}}(\mathbf{x}))))$.\\
        }
    }
    4.3. $T.\mathit{append}(t_d)$; $d \leftarrow m - i.$\\
    \uIf {$d > 0$}{
        4.4.1. $t_d \leftarrow \{\}$.\\
        \For{$j \leftarrow 1$; $j \leftarrow j + 1$; $j \leq t_{d+1}.\mathit{size}()$}{
            \For{$k \leftarrow 1$; $k \leftarrow k + 1$; $k \leq q^{d+1}_j.\mathit{size}()$}{
            4.4.2. $q^{\mathit{temp}} \leftarrow q^{d+1}_j.\mathit{remove}(index = k)$.\\
            \uIf{$q^{\mathit{temp}}$ \textbf{not in} $t_d$}{
                4.4.3. $t_d.\mathit{append}(q^{\mathit{temp}})$.\\
                4.4.4. $t_d.\mathit{connectParentChild}(q^{\mathit{temp}})$.\\
                }
            }
        }
        4.5. $T.\mathit{insert}(t_d, \mathit{index} = d)$.\\
    }
 }
 5. Output the grown tree $T$.\\
 \caption{Striking Method.}
 \label{the striking algorithm}
\end{algorithm}
\subsubsection{Computational Complexity and Scalability}
The Tree Growth algorithms are brute-force methods based on combinatorics, which may raise concerns about their consumed computational resources. Conceptually, pArborist generates all syntactically correct queries in Tree Growth and removes unqualified queries in Node Bypassing. However, this is only a generic procedure to maximize the generalization of pArborist. In practice, a natural optimization is to prevent the generation of unqualified queries that are syntactically correct but cannot retrieve any complex events from the data while growing trees, and thus, the computational complexity of Tree Growth algorithms for a given set of data of size $n$ is approximately $O(n)$. Note that the size $n$ does not refer to, e.g., the number of rows in datasets, but the number of complex events in a set of data if we study the data with set theory. Furthermore, because trees are grown upon a stable schema, they can typically be reused while scaling. If not, the grown trees can still form the basis of new trees, leading to an extension of existing trees instead of rebuilding new trees. The evaluation in Section \ref{evaluation} regarding efficiency and practicality indirectly confirms this conclusion by analyzing the CPU time consumed by pArborist. It shows that an Intel i7-13700 CPU and 32 GB memory is sufficient to provide state-of-the-art performance using pArborist within a reasonable amount of CPU time on public datasets.  

\subsection{Node Bypassing Algorithm} \label{NBA}
The algorithm aims to verify that the nodes in the constructed tree can reveal the same private information as the seed query. The verified nodes are the derived privacy-revealing queries, while the others are removed from the tree. This is performed in two steps:
\begin{enumerate}
    \item We analyze the correlation between each node and the seed query. We remove uncorrelated nodes and reconnect their parent and child nodes, i.e., bypass uncorrelated nodes.
    \item After bypassing the uncorrelated nodes, we continue to bypass the nodes that are conditionally independent of the seed. The condition for calculating conditional dependence can be any other node that remained in the tree. 
\end{enumerate}
The verification can be processed differently for databases or data streams, when given sufficient or insufficient data, i.e., when the uncertainty of the calculated statistics are sufficiently low or not. We summarize the solutions of these scenarios into a Node Bypassing algorithm with three versions: (1) The population approach is applied to databases and the warm-up of data streams if given sufficient data. (2) The bootstrap approach is applied to databases and the warm-up of data streams if given insufficient data. (3) The incremental approach is applied to the data streams after the warm-up. The warm-up refers to the preparation phase before a data stream application is deployed. We first present the general principles for verification that are shared among the three approaches. Then, we introduce the Node Bypassing algorithm and its three approaches.

\subsubsection{Verification of Correlation and Conditional Dependence}
Before presenting the two verification steps, we first explain the reason for selecting correlation and conditional dependence as the verification metrics. For correlation, if the outputs of two queries are not correlated, they are independent of each other. For a node that reveals the same private information as the seed node, its correlation with the seed is a necessary condition. However, correlation is not a sufficient condition to confirm that the node is privacy-revealing. For example, if it rains, the number of pedestrians on the street decreases, and the water level of a river increases. It appears that there is a correlation between the number of pedestrians and the water level, although they are conditionally independent when it rains. Without considering rain, the decrease in pedestrians does not cause or reveal the increase in water levels. One of these events cannot be used to infer the other. Therefore, conditional dependence is applied as an addition metric to correlation.

Regarding the metrics for Step (1), three existing state-of-the-art correlation coefficients are applied, i.e.,  Cohen's Kappa ($\kappa$) \cite{cohen1960coefficient}, Kendall's rank correlation coefficient ($\tau$) \cite{kendall1938new}, and intraclass correlation coefficient (ICC) \cite{koch2004intraclass}. By default, $\kappa$ is applied, but if the queries for verification have no predicates, that is, the queries retrieve all values of certain attributes, we may apply $\mathit{ICC}$ or $\tau$. In such cases, if all retrieved values are categorical and can be ranked, e.g., the grades of A, B, or C, $\mathit{\tau}$ is applied, and if the values cannot be ranked, $\kappa$ is still applied. For all other unspecified cases, $\mathit{ICC}$ is applied. 

To implement Step (1), we execute the seed query and a generated query on the given set of data. Their answers can be used to calculate the corresponding correlation coefficients. For example, if the generated query and the seed query have predicates, their resulting data tuples can be labeled with a binary category, indicating the presence or absence of a complex event. In such cases, $\kappa$ is calculated as the correlation coefficient. The correlation coefficients range from $-1$ to $1$ or $0$ to $1$. Although their ranges are different, larger absolute values of coefficients always correspond to stronger correlations by definition, and whether the correlation is indicated by a negative coefficient or a positive coefficient is not important for this work. Therefore, this paper considers that there is a significant correlation between the answers of the seed query and the generated query if the absolute values of correlation coefficients are greater than $\alpha = 0.5$. $\alpha$ is a tunable parameter. With setting $\alpha = 0$, any agreement between the answers of the seed query and the generated query is considered a significant correlation, while with $\alpha = 1$, only a full agreement is considered significant. 

Before verifying the conditional dependence in Step (2), we first eliminate the nodes that are not correlated with the seed. After the removal of uncorrelated nodes, we reconnect their parent nodes with their child nodes. Note that their corresponding depths are not changed, since they correspond to the numbers of A-F pairs. We reconstruct the connection between nodes because: (1) the child nodes of the removed node can remain in the tree until they are proven invalid, and (2) if more computational resources are available, we may continue to grow the tree with more nodes. 

Eliminating invalid nodes and reconstructing the tree significantly reduces the calculation efforts during Step (2). For a seed node $A$ and a candidate node $B$, they are always conditionally dependent if given a condition $C$ that is not correlated with $A$, since $P(A|C) = P(A) \neq P(A|B)=P(A|B,C) $ always holds.
\begin{definition}
    For events $\mathit{A}, \mathit{B},$ and $\mathit{C}$, $\mathit{A}$ and $\mathit{B}$ are conditionally dependent given $\mathit{C}$ if and only if $P(C) > 0$ and $P(A|B,C) \neq P(A|C)$.
\end{definition}
Based on the definition above, we verify in Step (2) the conditional dependence between each node remaining in the tree and the seed node, when given any other node in the tree as the condition. If a node is always conditionally dependent on the seed, we confirm that it is a valid privacy-revealing query regarding the given seed. We remove invalid nodes and reconstruct the tree as in Step (1).

The verification is completed after Steps (1) and (2). However, the reliability of the calculated correlation coefficients is highly sensitive to the amount of data used in the calculation. This indicates that the correlation verification performed in Step (1) can exhibit considerable uncertainty. Therefore, we introduce the \textbf{confidence interval} (CI) of the absolute values of these coefficients at a confidence level $95\%$ based on field experience \cite{du2009confidence} to analyze the uncertainty and confirm the calculated correlation. For the CI of the absolute value of a calculated coefficient, if its upper limit is lower than $0.5$, we conclude that, with a $95\%$ chance, the actual correlation is also lower than $0.5$. Similarly, if its lower limit is greater than $0.5$, the actual correlation is with a $95\%$ chance greater than $0.5$. In such cases, our conclusions are consistent regarding the calculated correlation and its uncertainty. However, if the lower limit of CI is less than $0.5$, while the upper limit is greater than $0.5$, the conclusions are inconsistent, indicating a great uncertainty in the calculation and that the given amount of data is insufficient. We clarify that the amount of data is \textbf{sufficient} for the verification of a node if the lower and upper limits of its CI agree with each other, i.e., both are greater or smaller than $0.5$. Otherwise, we consider that the given amount of data is \textbf{insufficient}.

\subsubsection{Node Bypassing Algorithm} \label{node bypassing algorithms}
The Node Bypassing algorithm comprises the solutions for both databases and data streams with sufficient or insufficient data and merges them into an integrated automatic algorithm, as presented in Algorithm \ref{the node bypassing algorithm}. We use databases as the basis of this algorithm and consider data streams as databases with (1) continuous updates and (2) requirements for delays regarding real-time processing. The algorithm is also capable of processing the updates of an originally static database. We assume that there is an initial set of data as a warm-up dataset for data stream scenarios. 

\begin{algorithm}
\SetAlgoLined
 1. $S_d \leftarrow \mathit{dataset}$; $T:=\{t_{i}, t_{i+1}, ..., t_{i+d}\}$; $t_i:=\{q^i_1, q^i_2, ..., q^i_{n_i}\}$.\\
 2. $t_0 \leftarrow \mathit{time}()$.\\
 \uIf{$!\mathit{newData}$}{
    \uIf{$\mathit{sufficiencyCheck}(S_d, T) == \mathit{True}$}{
    3.1.1. $T.\mathit{verify}(S_d$, \textbf{population}).\\
     }
     \uElseIf{$!= \mathit{simulatedData}$}{
        3.2.1. $T.\mathit{verify}(S_d.\mathit{append}(\mathit{simulatedData})$, \textbf{population}).\\
     }
     \Else{
        3.3.1. $T.\mathit{verify}(S_d.\mathit{bootstrap}()$, \textbf{bootstrap}).\\
     }
     3.4. $T^{\mathit{output}}_1 \leftarrow T.\mathit{reconstruct}()$.\\
 }
 4. $t  \leftarrow \mathit{time}() - t_0$.\\
 \Else{
    \uIf{$t \leq \mathit{permittedDelay}$}{
        5.1.1. $T.\mathit{update}(\mathit{newData}$, `\textbf{lazyUpdating}').\\
    }
    \Else{
        5.2.1. $T.\mathit{update}(\mathit{newData}$, `\textbf{earlyAbandoning}')\\
    }
    5.3. $T^{\mathit{output}}_2 \leftarrow T.\mathit{reconstruct}$().\\
 }
 \caption{Node Bypassing Algorithm.}
 \label{the node bypassing algorithm}
\end{algorithm}

In Steps 1 and 2 of Algorithm \ref{the node bypassing algorithm}, we prepare the constructed tree and the given set of data, while in Step 3, we perform the calculation of the three correlation coefficients. If the lower and upper limits of their corresponding CIs agree on each other, we regard the amount of data as sufficient, otherwise insufficient. For the experiments described in this paper, 700 data rows can usually be considered as a sufficient amount of data. For sufficient data, we apply the \textbf{population approach} (Step 3.1.1), which directly calculates the coefficients for any given node using all available data. Calculations of privacy concerns can be executed on a trusted device. The coefficients and their CIs do not contain specific information or data and are not private. These results can be processed on any device and directly used to remove invalid nodes. After the reconstruction of the tree, the verification of conditional dependence follows the procedure proposed before. Eventually, the Node Bypassing algorithm produces a completed tree, in which all nodes reveal the same private information as the seed.

If there is insufficient data, we consider two scenarios (Steps 3.2 and 3.3). In some cases, the underlying model of a given dataset is known. We can generate synthetic data through simulation based on the model until its amount reaches the requirement (Steps 3.2.1 and 3.2.2). For example, medical treatments for an individual can be predicted by a health professional if given the corresponding health status. When data on these treatments are insufficient, it is possible to generate more synthetic data by simulation based on medical expertise. In such cases, the population approach is applied.

If such models are not available, we apply the \textbf{bootstrap approach} (Steps 3.3.1 and 3.3.2), which generates data by statistical bootstrapping, i.e., randomly sampling with replacement. We sample from the given set of data a new dataset of the same size and iteratively perform this procedure to satisfy the requirement for calculating CIs. The corresponding coefficients, CIs, and conditional dependence can then be verified. For example, assuming that we have generated $n = 2000$ sets of data of the same size as the original dataset, for any coefficient $C$, $n$ coefficients $C_i$ can be calculated. Their average $\bar{C}$ is the estimated expectations of $C$. We sort all coefficients in an ordered list $(C_1, C_2, ..., C_{2000})$, and $[C_{\frac{95\%}{2}n}, C_{n - \frac{95\%}{2}n}]$, i.e., $[C_{50}, C_{1950}]$, is the $95\%$ two-side confidence interval. After we complete a tree with either sufficient or insufficient data, we deliver all nodes to data producers as derived privacy-revealing queries (Step 3.4). This approach does not guarantee sufficiently narrow CIs. If and only if the upper limit of a CI is higher than a configurable threshold, which is $0.5$ in this paper, we consider the corresponding query as privacy-revealing.

In Step 5, we present the \textbf{incremental approach} to update a completed tree when a database is updated or more data are streamed. We regard the update of a database as a specific case of data streams and take the data stream scenario as the basis to present the incremental approach. Steps 1-4 are the warm-up for data stream scenarios. We assume that each attribute in the given data has a unique semantic representation and that there are no attribute conflicts between the initial data and the incoming data. The core challenge of the incremental approach is that the status of the nodes can change according to the new data in a data stream. The invalid nodes may become valid, or vice versa. An intuitive approach is to maintain all candidate nodes generated by Tree Growth algorithms, update their status, and reconstruct trees if receiving new data. This approach keeps all candidates and performs frequent updates, leading to a heavy consumption of computational resources. Therefore, this approach rarely meets the real-time processing requirement for data stream applications. There are two solutions to ease this shortcoming: (1) the \textbf{lazy-updating} solution (Step 5.1.1) reduces the frequency of updating the status of maintained nodes, and (2) the \textbf{early-abandoning} solution (Step 5.2.1) reduces the number of nodes that are kept.

Early-abandoning permanently removes a candidate node if the upper limit of its $95\%$ CI is smaller than $0.5$. If updating the remaining nodes is still a burden for real-time processing, we remove more nodes based on a smaller CI, e.g., $90\%$ CI. We keep reducing the size of CIs and removing nodes until the number of nodes allows for real-time processing. For extreme cases in which even the smallest CI cannot meet the real-time requirement, we stop updating the nodes. The updates of nodes for early-abandoning is performed by Bayesian inference, since the incoming data are usually of small size and the update occurs frequently. The lazy-updating approach is performed asynchronously for real-time data stream applications. If there are sufficient computational resources and data in addition to those used for real-time processing, we can asynchronously re-verify all remaining nodes and update the tree. The update follows the same procedure applied to the initial set of data, and the new tree replaces the tree in use once the update is completed.

Early-abandoning may permanently omit certain critical privacy-revealing complex events regardless of the amount of new data, while lazy-updating endangers the performance of pArborist during real-time processing. Therefore, a combination of both can be a superior solution. Based on the latest verification, we can cache the approximate computing time $\hat{t}$. When the data stream application is running in real time, the early-abandoning approach is applied by default. If the allowed delay is longer than $\hat{t}$, i.e., if there are available computational resources, lazy-updating can be executed asynchronously with the early-abandoning approach. We conclude that this combination is the complete incremental approach and that database updates can be processed in the same way. The updated tree can be reconstructed and delivered to data producers if they require this (Step 5.3). We notice that the calculation of lazy-updating can be divided into multiple smaller partitions, so that each partition of calculation can be executed simultaneously when there are available computational resources in real-time processing. However, it is usually challenging to measure the amount of available computational resources in such cases, and thus, this division is not included in pArborist or evaluated in this paper. 

\section{Evaluation} \label{evaluation}
We evaluate pArborist in terms of its (1) performance, i.e., the capability of deriving privacy-revealing queries and the verification of performance with the state-of-the-art approach FQID \cite{podlesny2023quasi}, (2) efficiency, i.e., the amount of required computational resource, and (3) practicality, i.e., the burden put onto data producers and consumers.
\begin{table*}
\small
\centering
\begin{tabular}{|c||c|c|c|} 
 \hline
 Initial input seed & Confirmed seed & Examples of derived results\\ 
 \hline \hline
 \multirow{5}{12em}{GDPR Seed: $\{(\mathit{name}, f_{\mathit{eq}}(\mathit{Bob}))$, $(\mathit{venue}, f_{\mathit{eq}}(H_1))\}$} & \multirow{3}{11em}{$\{(\mathit{name}, f_{\mathit{eq}}(\mathit{Bob}))$, $(\mathit{venue}, f_{\mathit{eq}}(H_1))\}$} & $\{(\mathit{name}, f_{\mathit{eq}}(\mathit{Bob}))$, $(\mathit{long}, f_{\mathit{eq}}(x_1))$, $(\mathit{lati}, f_{\mathit{eq}}(y_1))$\} 
 \\ \cline{3-3}
 &  &$\{(\mathit{name}, f_{\mathit{eq}}(\mathit{Bob}))$, $(\mathit{brand}, f_{\mathit{eq}}(B))$, $(\mathit{country}, f_{\mathit{eq}}(C))$\} 
 \\ \cline{3-3} 
 &  &$\{(\mathit{name}, f_{\mathit{eq}}(\mathit{David}))$, $(\mathit{ALL}, f_{\mathit{eqjoin}}(\mathit{name}_{\mathit{checkin}}, \mathit{name}_{\mathit{Bobfriends}})$, $(\mathit{venue}, f_{\mathit{eq}}(H_1))$\} 
 \\ \cline{2-3}
 & \multirow{2}{11em}{$\{(\mathit{name}, f_{\mathit{eq}}(\mathit{Bob}))$, $(\mathit{venue}, f_{\mathit{eq}}(H_2))\}$} & $\{(\mathit{name}, f_{\mathit{eq}}(Bob))$, $(\mathit{long}, f_{\mathit{eq}}(x_2))$, $(\mathit{lati}, f_{\mathit{eq}}(y_2))$\} 
 \\ \cline{3-3}
 & & $\{(\mathit{name}, f_{\mathit{eq}}(\mathit{David}))$, $(\mathit{date}, f_{\mathit{within}}(04122359, 04132359))$\} \\ 
 \hline 
 \multirow{6}{12em}{Manual Seed: $\{(\mathit{uID}, f_{\mathit{eq}}(\mathit{ANY})$, $(\mathit{location}, f_{\mathit{eq}}(\mathit{place001}))$, $(\mathit{age}, f_{\mathit{within}}(18, 21))\}$} & \multirow{3}{12em}{$\{(\mathit{uID}, f_{\mathit{eq}}(\mathit{ANY})$, $(\mathit{location}, f_{\mathit{eq}}(\mathit{place001}))$, $(\mathit{age}, f_{\mathit{within}}(18, 21))\}$} & \multirow{1}{*}{$\{(\mathit{uID}, f_{\mathit{eq}}(\mathit{ANY})$, $(\mathit{location}, f_{\mathit{eq}}(\mathit{place001}))$, $(\mathit{drinks}, f_{\mathit{eq}}(0))\}$}  \rule[-2ex]{0pt}{5ex}
 \\ \cline{3-3}
 & & \multirow{1}{*}{$\{(\mathit{uID}, f_{\mathit{eq}}(\mathit{ANY})$, $(\mathit{language}, f_{\mathit{eq}}(\mathit{Eng}))$, $(\mathit{drinks}, f_{\mathit{eq}}(0))\}$} \rule[-2ex]{0pt}{5ex}
 \\ \cline{2-3}
 & \multirow{3}{12em}{$\{(\mathit{uID}, f_{\mathit{eq}}(\mathit{ANY})$, $(\mathit{location}, f_{\mathit{eq}}(\mathit{place006}))$, $(\mathit{age}, f_{\mathit{within}}(18, 21))\}$} & 
 \\& & \multirow{1}{*}{$\{(\mathit{uID}, f_{\mathit{eq}}(\mathit{ANY})$, $(\mathit{location}, f_{\mathit{eq}}(\mathit{place006}))$, $(\mathit{drinks}, f_{\mathit{eq}}(0))\}$}  \rule[0ex]{0pt}{0ex}
 \\ & & \\ 
 \hline 
\end{tabular}
\caption{Examples\protect\footnotemark of derived results}
\label{examplesOfSeeds}
\vspace{-16pt}
\end{table*}

\subsection{Evaluation Metrics} \label{EM}
We present and reason for the selected evaluation metrics for performance, efficiency, and practicality. Performance is measured by recall and precision. Efficiency is measured by the usage of computational resources, i.e., CPU time, when deriving privacy-revealing queries. Practicality is indirectly measured by the number of seeds required from data producers. Performance and practicality are evaluated for the entire pArborist, while efficiency is only evaluated for the derivation phase because the CPU time used in the seeding phase is insignificant, i.e., on average less than $1$ ms to generate the seed queries for the dataset used in the experiment. 

To measure performance, we apply recall and precision of the derived privacy-revealing queries. Although the number of derived privacy-revealing queries $q^d$ can be a more intuitive metric, this number is highly dependent on the given data and lacks generalization. The precision can be directly calculated in experiments, which requires the number of correctly derived privacy-revealing queries and the number of all derived queries. In contrast, it is challenging to calculate recall because there are infinite privacy-revealing queries. Since recall can only be produced if the experiments are of finite scope, we set a limit on the maximum CPU time.

To measure efficiency, we illustrate the CPU time consumed by each depth of a tree in the derivation phase. For example, for a tree of a total depth $n$, we first perform the derivation phase only at depth $1$, including the Tree Growth and Node Bypassing steps, and illustrate the CPU time consumed. Subsequently, we continue to perform the derivation phase at depth $2$ based on the results at depth $1$ and illustrate the additional time consumed only for depth $2$. We continue this procedure until depth $n$, when we take the grown tree of depth $n-1$ as a basis and illustrate the consumed CPU time only at depth $n$.

The practicality can be evaluated from two perspectives: (1) Since data producers may only be required to provide seed queries and the remaining steps of pArborist are automatic, the practicality can be analyzed based on the number of seed queries required from data producers. (2) The evaluation of performance and efficiency also provides insight into practicality, e.g., the CPU time used.

\subsection{Experiments}
We present the datasets and the design of experiments that evaluate the recall, precision, efficiency, and practicality of pArborist. The experiments were carried out using Python and C++ on a Windows 11 platform with an Intel i7-13700 CPU and 32 GB memory.\footnotetext{The actual values of data tuples are anonymized. We endow them with practical names and values only for better demonstration.} 

\subsubsection{Datasets}
The experiments are conducted based on three real-world datasets, i.e., the \textbf{Election} dataset \cite{sabuncu10usa}, the \textbf{Check-in} dataset \cite{yang2019revisiting,yang2020lbsn2vec++}, and the \textbf{APP} dataset \cite{articleocc}. The Election dataset contains over 20 million records of the political opinions and profiles of voters during the 2020 US presidential election. The Check-in dataset contains over 10 million records of check-ins for over $10000$ hotels and restaurants, including the time of check-in, names of customers, names of venues, reviews, etc. APP contains over $50000$ personal records for APP users, including gender, unique ID, locations, etc. The Election dataset is applied by FQID in its original research, which is selected as a comparative approach to demonstrate the effectiveness of pArborist. To the best of our knowledge, Election is the only labeled public dataset that satisfies the requirements for our evaluation. Therefore, the other datasets used in the related works are not utilized in the evaluation. In addition, the Check-in dataset is used to evaluate pArborist in finding privacy-revealing queries according to GDPR. The seed queries of this experiment are formulated based on GDPR regulations, and no additional seeds are provided by our simulated data producers. APP consists of more complex data tuples than Check-in and is used to evaluate pArborist for complex use cases and queries. All seed queries applied for APP and Election are formulated by our simulated data producers in the seeding phase. We classify the seeds used for Check-in and APP as \textbf{GDPR seeds} and \textbf{manual seeds}. Examples of these seeds and their derived results are listed in Table \ref{examplesOfSeeds}.

The labels for privacy-revealing complex events in the Election dataset are public, and Check-in and APP datasets are manually labeled \cite{parborist}. In detail, we manually perform a more precise verification than the automatic algorithm of pArborist based on the tree grown by pArborist, according to (1) statistics, (2) external information provided by the owners of datasets, and (3) our privacy expertise. For example, for the Check-in dataset, GPS records reveal the same private information as the names of hotels. In the most extreme cases, there may exist only one data tuple that supports this correlation between GPS records and the names of hotels. This leads to a great uncertainty in calculating correlation coefficients. In such cases, given a seed query that retrieves GPS locations, pArborist cannot confirm with sufficient certainty that the names of hotels are also privacy-revealing. However, with our privacy expertise, we can manually label the names of hotels as privacy-revealing. The labeling process is performed in a privacy-conscious way, that is, if there is uncertainty whether a query reveals private information or not, the query is labeled as privacy-revealing.

\subsubsection{Design of Experiments}
We design experiments based on the presented evaluation metrics. To evaluate efficiency, two experiments were conducted on the Check-in and APP datasets, measuring the CPU time used in the derivation phase of pArborist. To evaluate performance, four experiments were conducted on the Check-in and APP datasets, which measure the recall and precision w.r.t. different numbers of seeds or depths of trees (different complexities of queries). We evaluate most fundamental operators, such as the basic predicates, Join clauses, aggregations with GROUP BY clauses, and UNION operators. Their corresponding complexities regarding A-F pairs and the depth of trees are discussed in Section \ref{FofS}. The results of these experiments are also analyzed to evaluate practicality. During experiments, the default number of seeds is six and the default depth of trees is eight. The maximum computing time is $30$ minutes and no experiment reached the limit. These default values are applied because they have the minimum influence on performance and efficiency. In addition, two experiments were conducted on the Election dataset to illustrate the effectiveness of pArborist compared to the state-of-the-art approach FQID w.r.t. recall and precision. pArborist identifies arbitrary privacy-revealing complex events, including quasi-identifiers, whereas FQID is limited to quasi-identifiers only, and thus cannot perform all tasks designed for pArborist. To enable a meaningful comparison, we simplify the tasks in this experiment and limit pArborist to quasi-identifiers by regarding attributes as events, because a comparison w.r.t. privacy-revealing complex events is meaningless (the recall of pArborist can exceed that of FQID by more than $60\%$).

For each experiment, both data stream and database scenarios are simulated. In data stream scenarios, $20\%$ data are provided to pArborist for its warm-up, and the remaining data are subsequently streamed with a throughput of approximately $10000$ data rows per second. In such cases, the incremental approach is applied. In database scenarios, both datasets contain sufficient data for pArborist, and the population approach is applied. We also aim to evaluate pArborist when given insufficient data and when the bootstrap approach is applied. Therefore, we randomly select $15\%$ data from each dataset to simulate a database scenario with insufficient data. We summarize that three scenarios are simulated: (1) A data stream scenario with the incremental approach. (2) A database scenario with sufficient data and the population approach. (3) A database scenario with insufficient data and the bootstrap approach.

Furthermore, since we are interested in evaluating the influence brought by the insufficiency of data and the effectiveness of bootstrap, we also evaluate the performance of the population approach for the database scenario with insufficient data, although this is not an application of pArborist, and hence, this scenario is not considered when evaluating efficiency.

\subsection{Results and Analysis}\label{results and disscu}
The results of experiments with the Check-in and APP datasets show recall over $90\%$\footnote{Since privacy-revealing queries are manually labeled in a privacy-conscious way, the recall in experiments is approximately $5\%$ lower than the actual recall, i.e., approximately two additional false negatives that pAborist fails to derive.} and precision over $93\%$ for the default configuration of pArborist. On average, pArborist leads to a computation time of approximately $500$ ms for database scenarios, while $1.3$ ms for data stream scenarios after a $920$ ms warm-up. The precision and recall have reached $90\%$ for more than six seeds and remain stable regardless of the complexity of the queries. We further detail and analyze the results w.r.t. performance, efficiency, and practicality.
\begin{figure*}
\centering
    \subfloat[\label{checkinb}]{\includegraphics[width=0.245\textwidth]{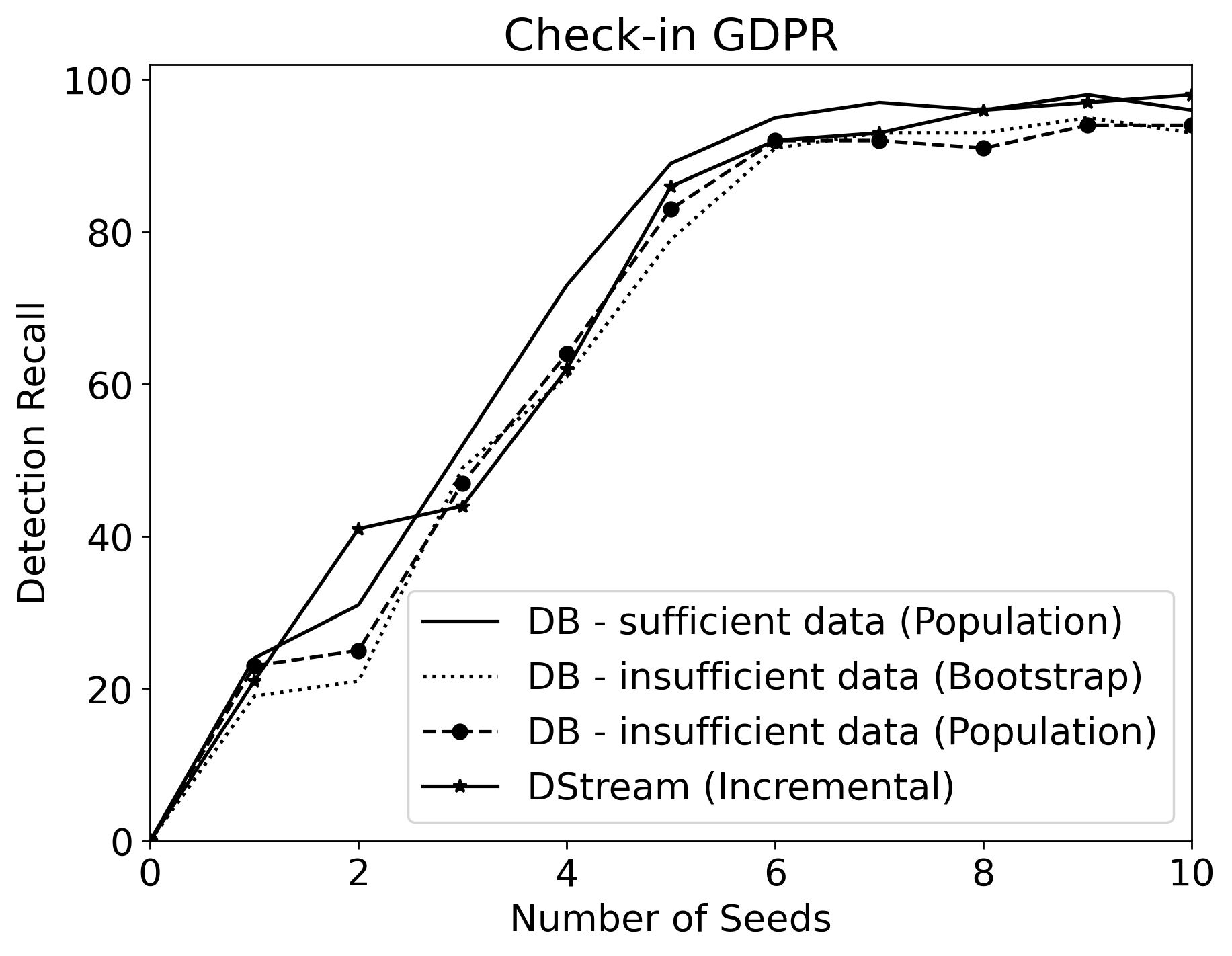}}%
    \subfloat[\label{appb}]{\includegraphics[width=0.245\textwidth]{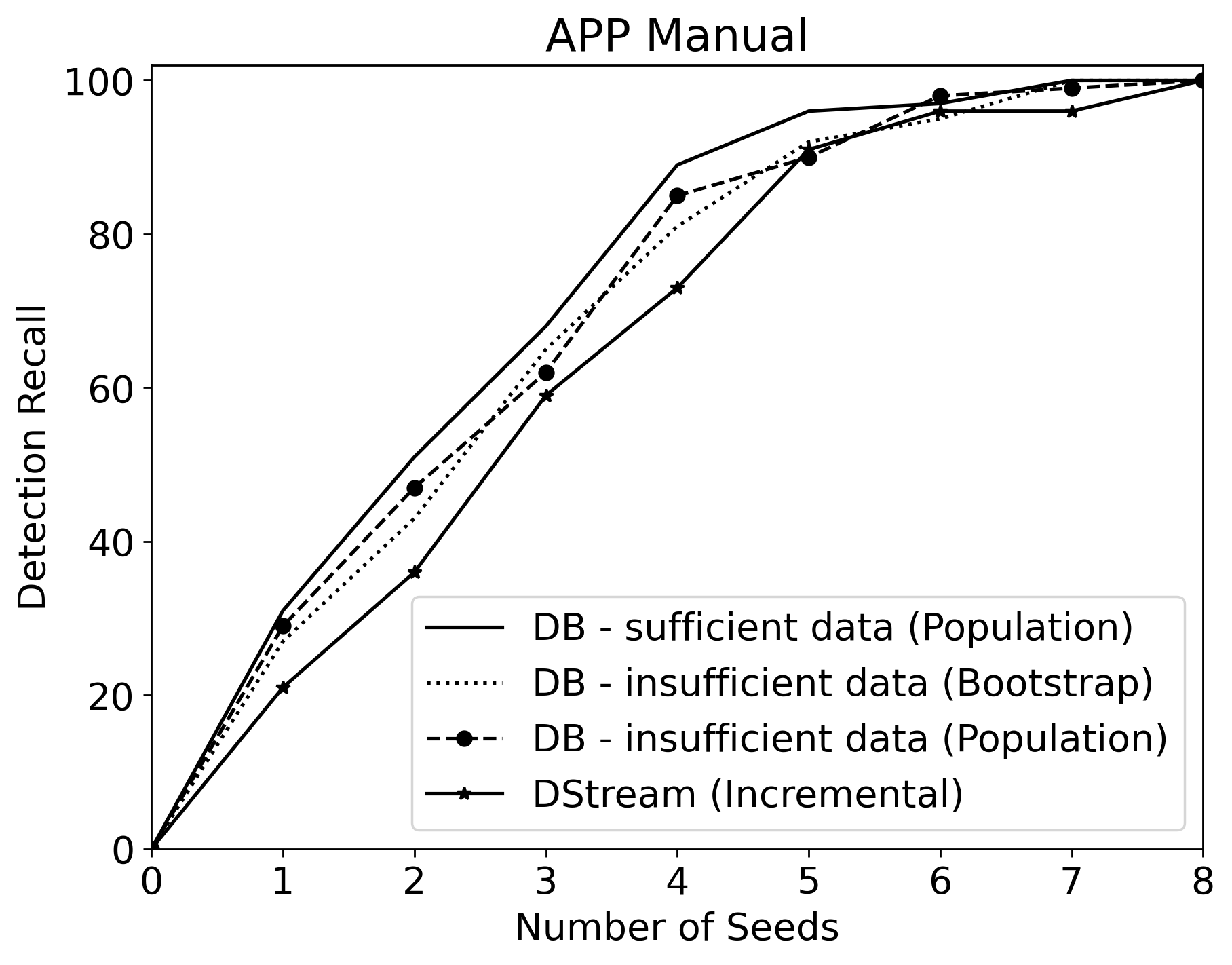}}%
    \subfloat[\label{checkind}]{\includegraphics[width=0.245\textwidth]{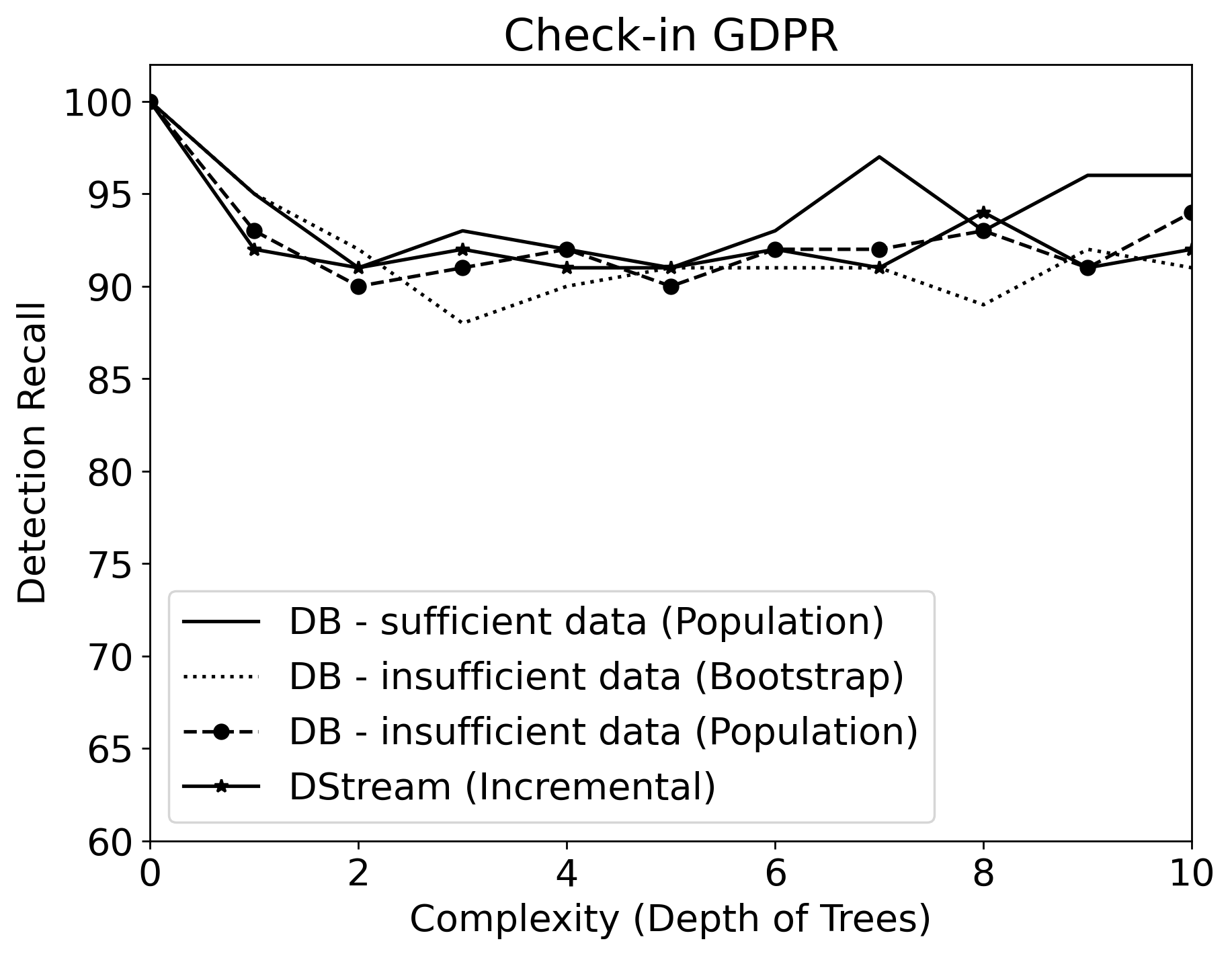}}%
    \subfloat[\label{appd}]{\includegraphics[width=0.245\textwidth]{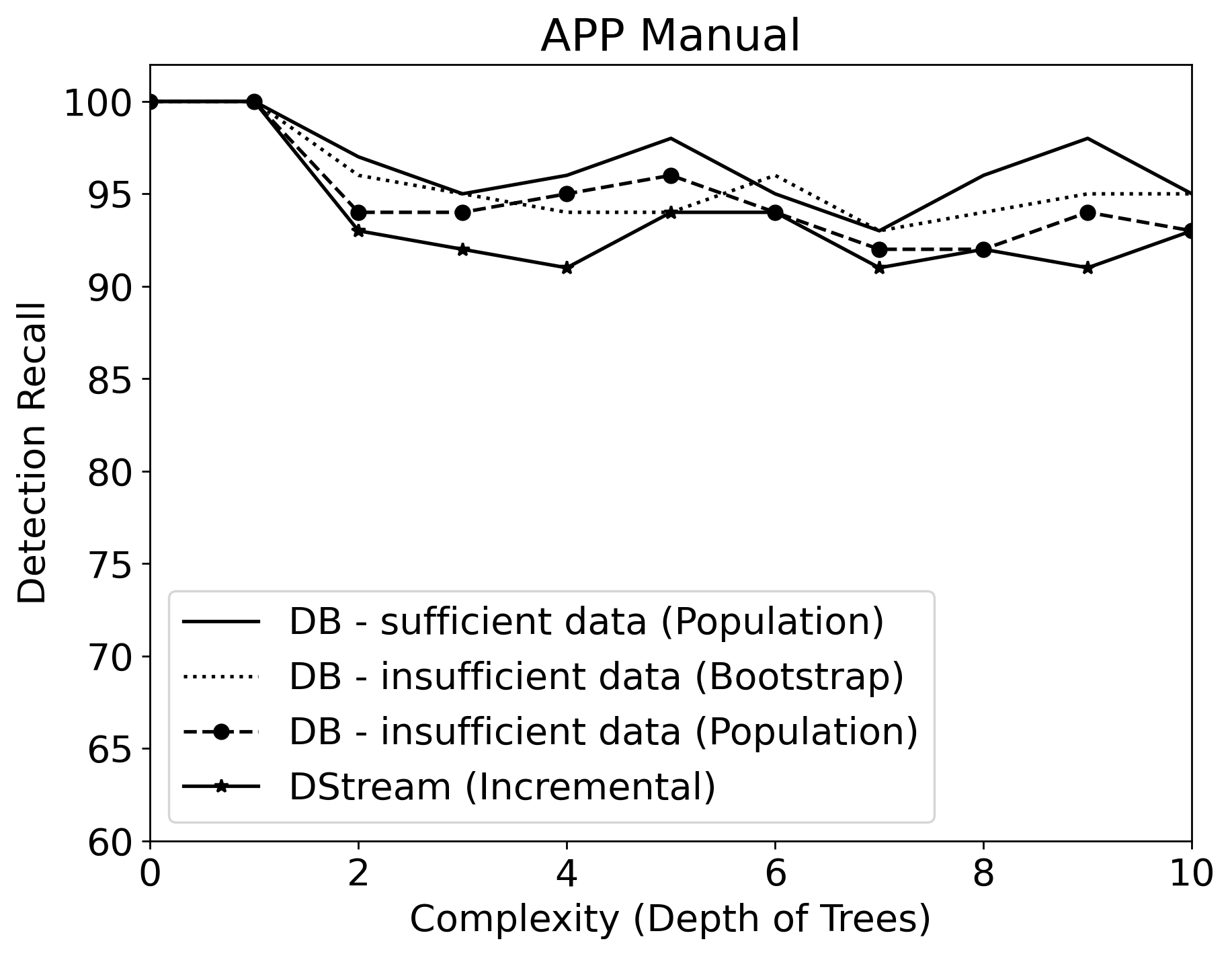}}%
\vspace{-7pt}
\caption{Recall on the Check-in Dataset and the APP dataset.}
\label{figure_recall}
\vspace{-15pt}
\end{figure*}
\begin{figure*}
\centering
    \subfloat[\label{checkinc}]{\includegraphics[width=0.245\textwidth]{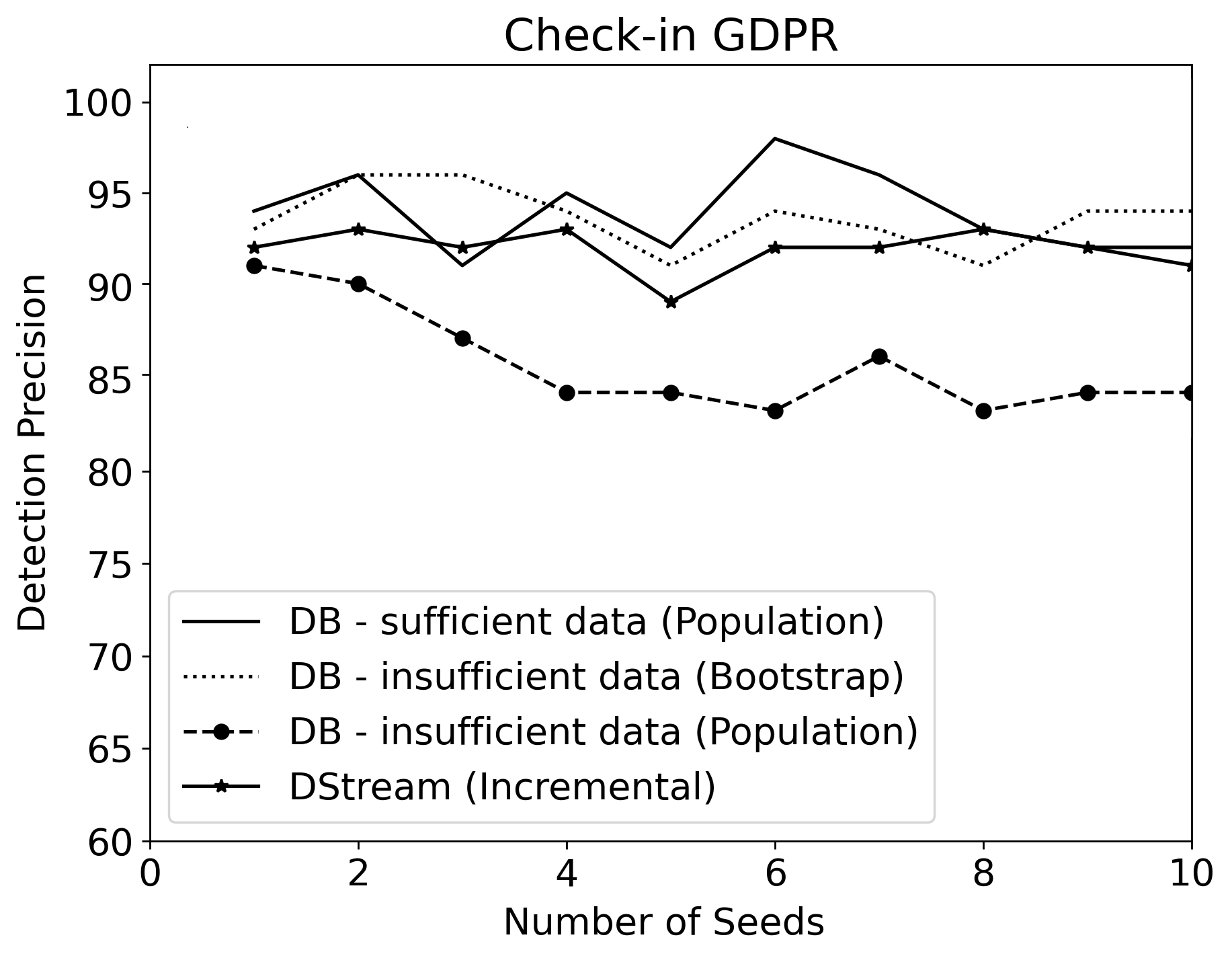}}%
    \subfloat[\label{appc}]{\includegraphics[width=0.245\textwidth]{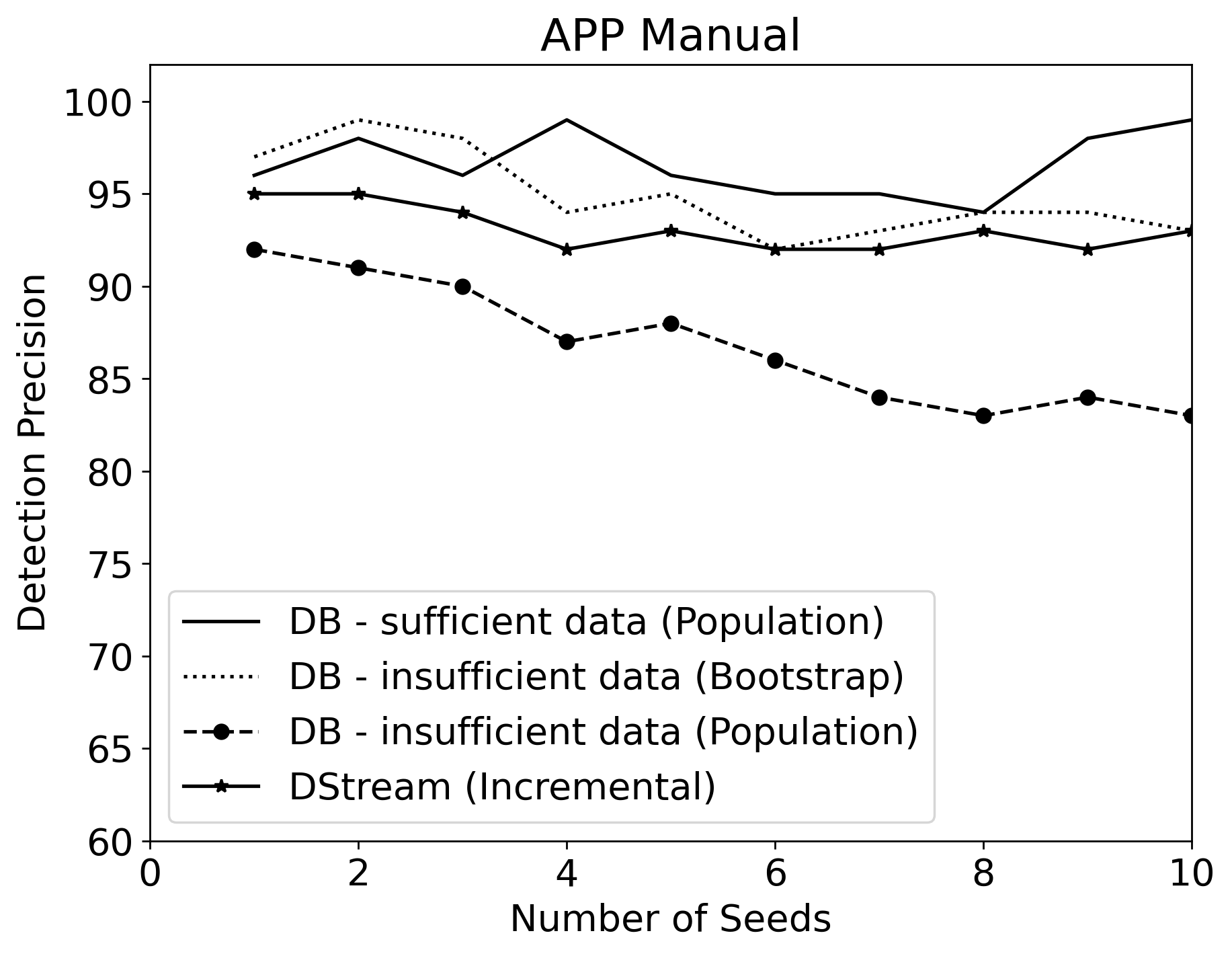}}%
    \subfloat[\label{checkine}]{\includegraphics[width=0.245\textwidth]{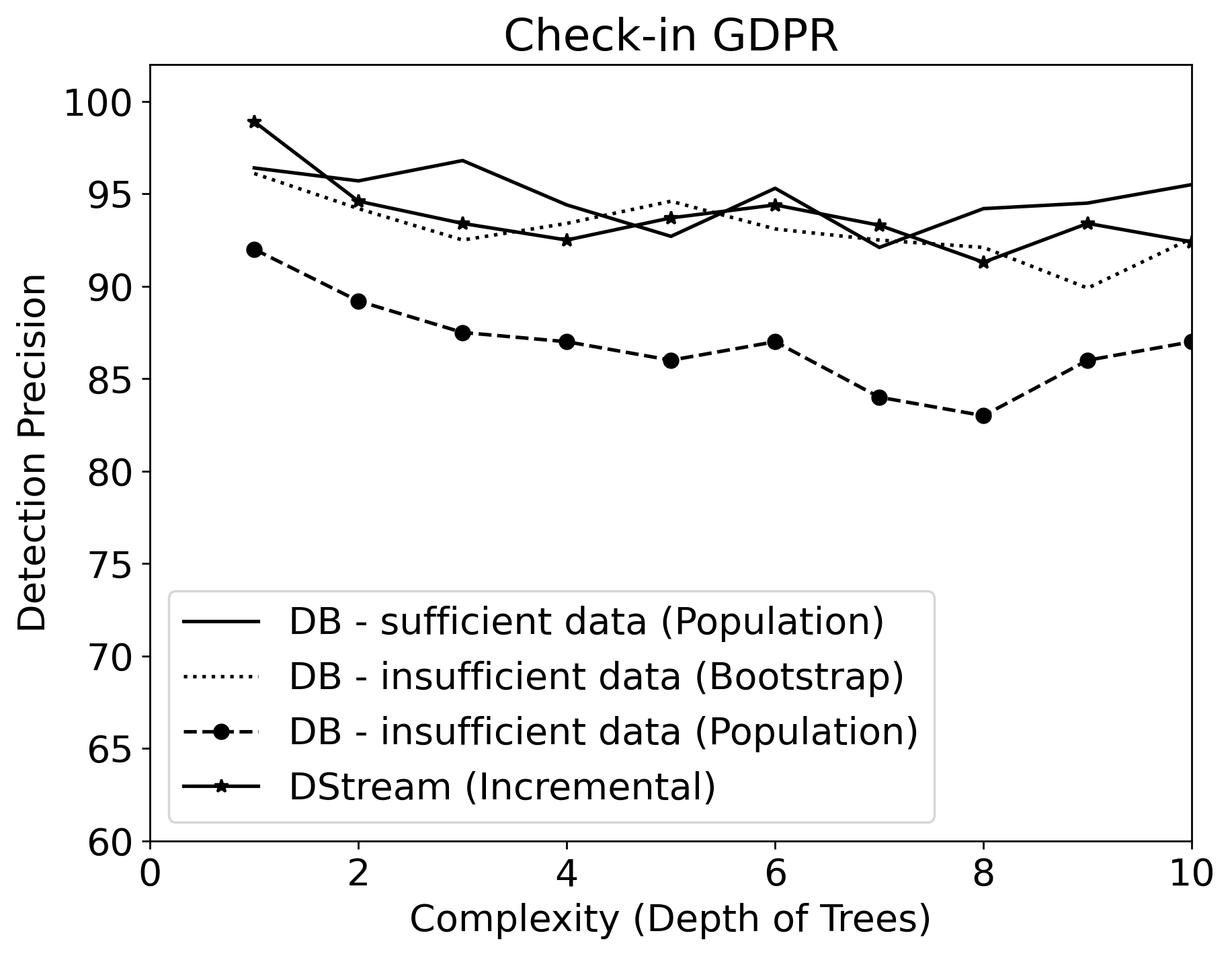}}%
    \subfloat[\label{appe}]{\includegraphics[width=0.245\textwidth]{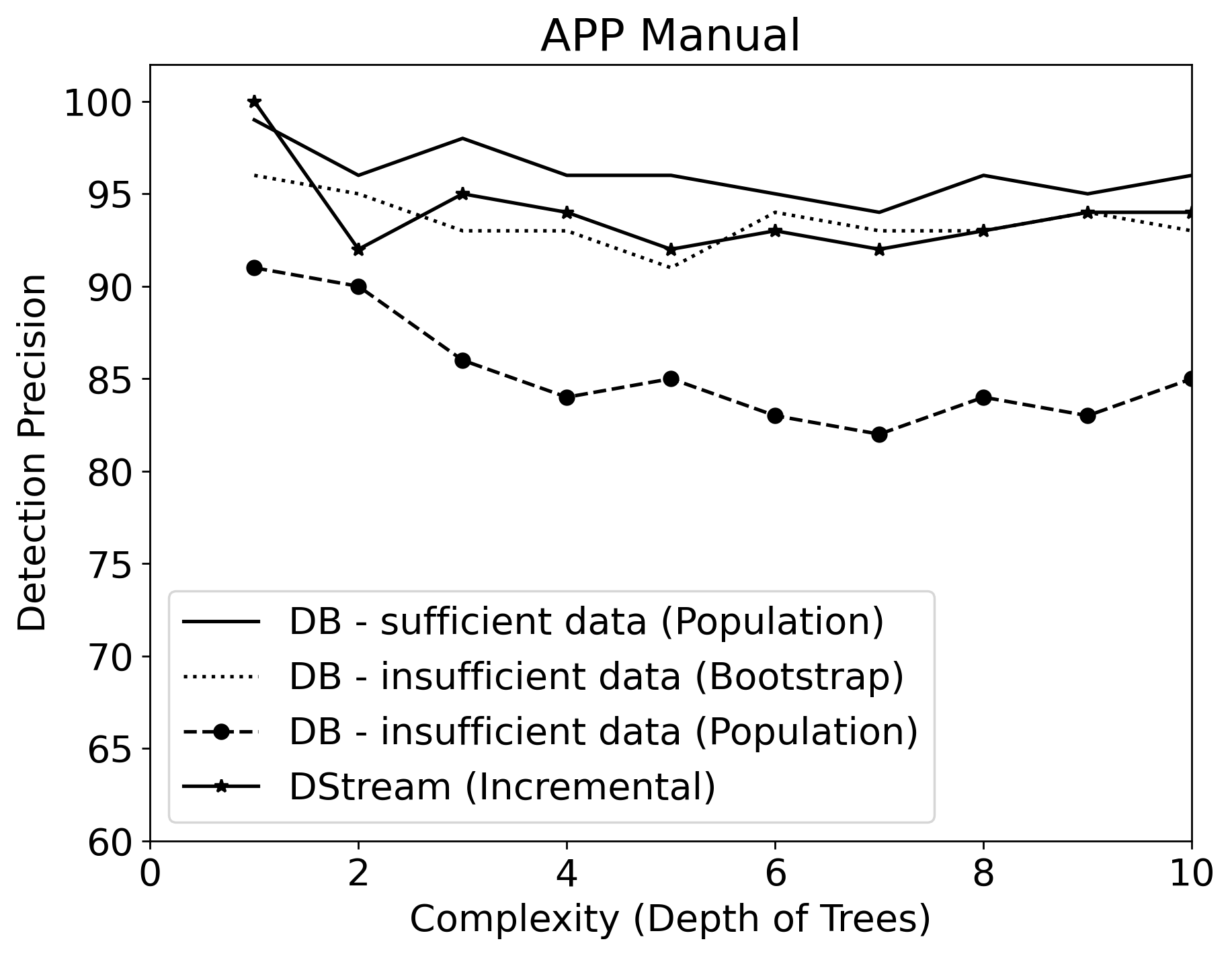}}%
\vspace{-7pt}
\caption{Precision on the Check-in Dataset and the APP dataset.}
\label{figure_precision}
\vspace{-15pt}
\end{figure*}
\begin{figure*}
\centering
    \subfloat[\label{recalla}]{\includegraphics[width=0.24\textwidth]{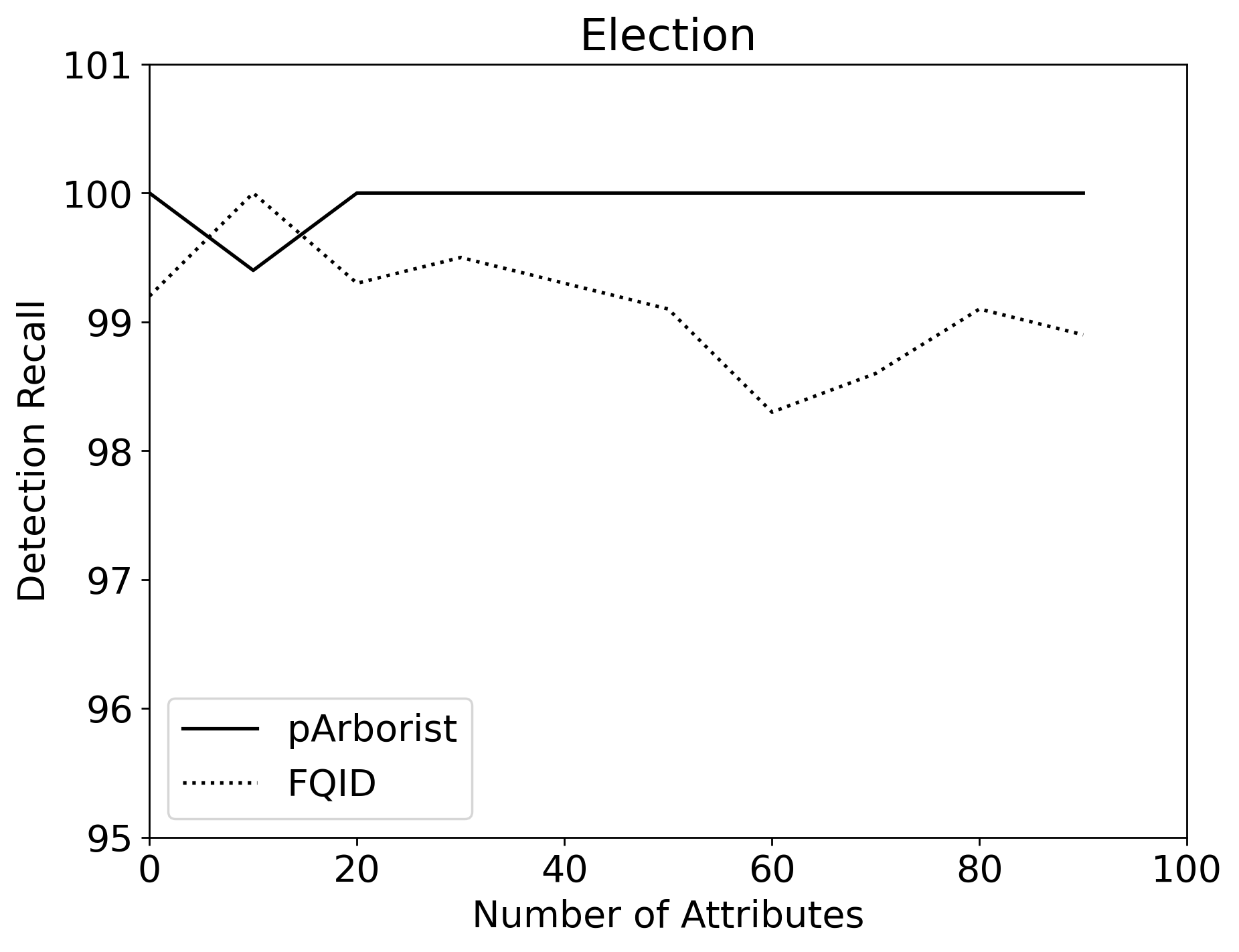}}%
    \subfloat[\label{precisiona}]{\includegraphics[width=0.245\textwidth]{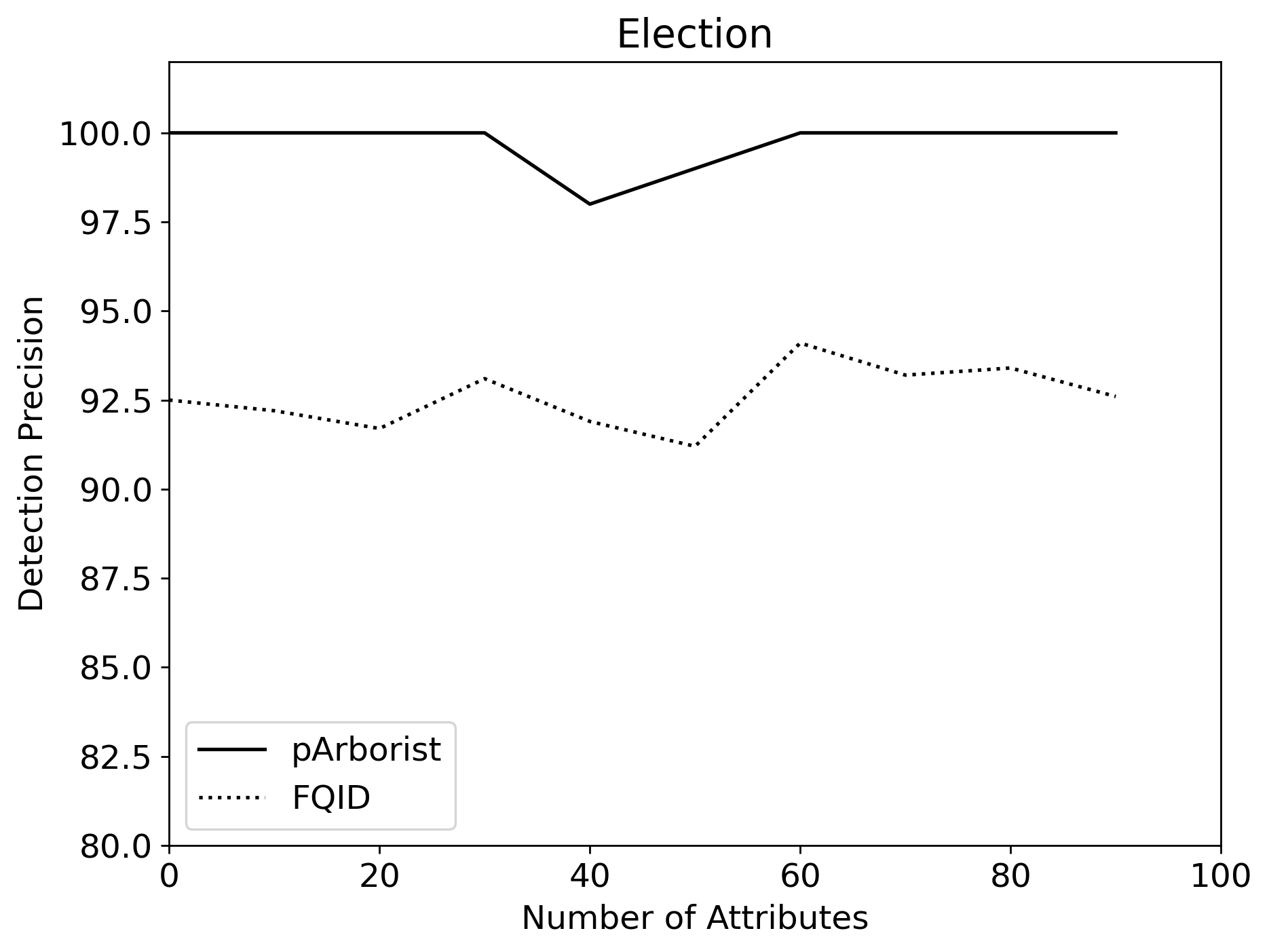}}%
    \subfloat[\label{checkina}]{\includegraphics[width=0.23\textwidth]{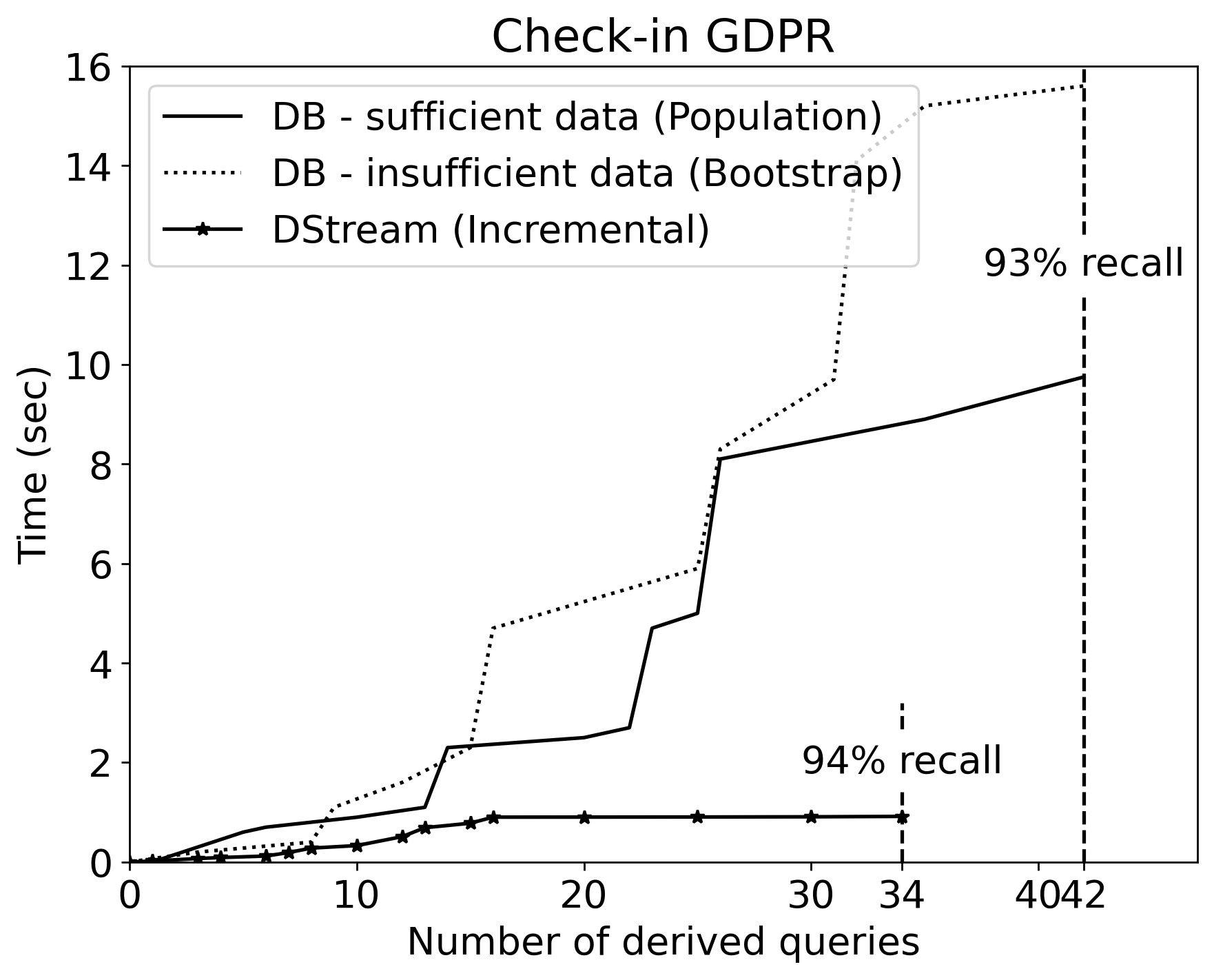}}%
    \subfloat[\label{appa}]{\includegraphics[width=0.24\textwidth]{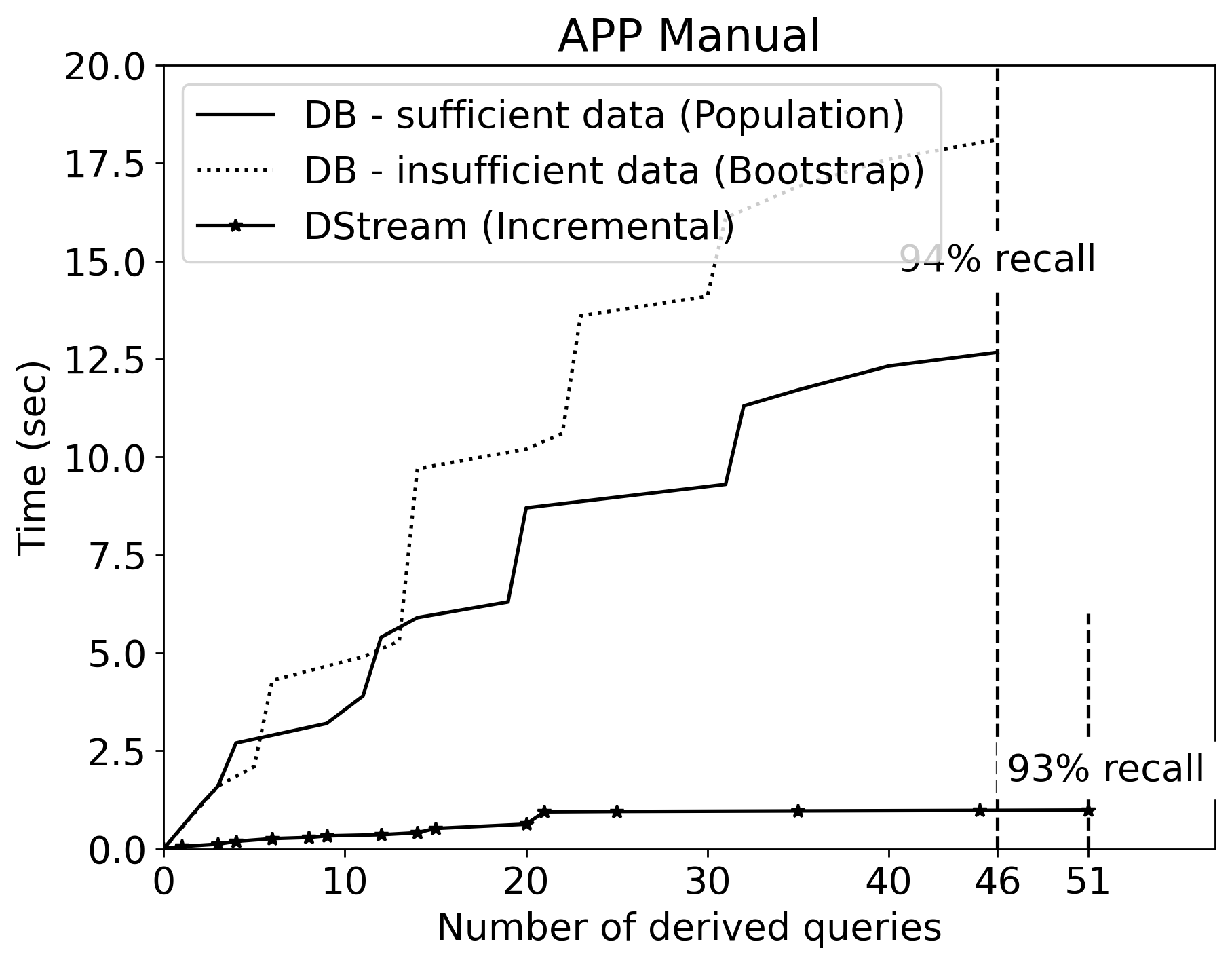}}%
\vspace{-7pt}
\caption{Recall and precisions of FQID and pArborist with the Election dataset. Efficiency with the Check-in and APP datasets.}
\label{figure_elec}
\end{figure*}

\subsubsection{Performance}
Figure \ref{figure_recall} illustrates the recall on the Check-in and APP datasets, while Figure \ref{figure_precision} illustrates the precision. Both are studies with respect to the number of seeds and the complexity of queries. Figure \ref{figure_elec} compares the recall and precision of FQID and pArborist on the Election dataset. Note that the recall and precision in the figures are magnified by expanding the y-axis to illustrate more details. A difference of $3\%$ corresponds to approximately only one or two falsely detected or omitted privacy-revealing queries, but is visualized by a rather steep line in the figures. 

Regarding recall, Figure \ref{checkinb} illustrates $95\%$ recall for databases with sufficient data for more than six seeds, $91\%$ recall for insufficient data, and $92\%$ recall for data streams, with the Check-in dataset. Figure \ref{appb} illustrates approximately $2\%$ higher recall with the \textbf{APP} dataset, reaching $97\%$ recall for databases with sufficient data, $95\%$ recall for insufficient data, and $96\%$ recall for data streams. Figures \ref{checkind} and \ref{appd} show approximately $90\%$ recall, regardless of the complexity of the queries. The recall remains stable for Check-in dataset and slightly decreases for APP dataset.

Analyzing the results illustrated by Figures \ref{checkinb} and \ref{appb}, we notice that the recall increases with the number of seeds. This matches our expectation, since pArborist can confirm more privacy-revealing queries if more private information is specified by seed queries. Figure \ref{checkinb} shows that the increase of recall brought by different seeds can significantly vary for the Check-in dataset, specifically when the number of seeds ranges from one to three. For the APP dataset, the increase in recall is smoother w.r.t. the increase of seeds. This indicates that each manual seed applied to the APP dataset has similar effects in helping derive privacy-revealing queries, while each GDPR seed may have significantly different contributions. Analyzing the results shown in Figures \ref{checkind} and \ref{appd}, we notice that the difference in recall for different datasets w.r.t. complexities is caused by the difference in their applied seeds. Since GDPR regulations usually correspond to simpler complex events, GDPR seeds are simpler than manual seeds. Therefore, the private information specified by GDPR seeds can mostly be revealed by simpler queries. In such cases, increasing complexity has less influence in deriving more privacy-revealing queries. In contrast, as manual seeds usually consist of more predicates, their specified private information is rarely revealed by simpler queries, and thus, queries of higher complexity have more influence on recall. 

Regarding precision, Figure \ref{figure_precision} shows that it remains stable and is usually higher than $90\%$. The average precision for the APP dataset is $3\%$ higher than that for the Check-in dataset. The figure indicates that applying the population approach to databases with insufficient data leads to significantly lower precision, i.e., $10\%$ on average, than applying the bootstrap approach following pArborist. The analysis of the results indicates that the precision remains stable because the usage of CIs eliminates a significant number of false negatives in advance. The bootstrap approach produces significantly better precision for insufficient data, because it produces CIs of better quality than the population approach. 

Regarding the results illustrated in Figure \ref{figure_elec}, pArborist performs significantly better than FQID w.r.t. recall and precision, with both being close to $100\%$. Analyzing the results, we notice that pArborist has a rather larger advantage in precision, while a smaller but still significant advantage in recall. pArborist succeeds in detecting more privacy-revealing queries, because FQID is proposed to detect quasi-identifiers that are formed only by attributes. In such cases, if only some specific values of an attribute are private, this attribute is not necessarily labeled as private by FQID, while pArborist can identify such scenarios and precisely label this type of data, since it is designed to detect privacy-revealing complex events. The larger advantage in precision is the result of the additional verification of privacy-revealing queries conducted by pArborist based on CIs, while FQID only takes recall as its performance metrics.

The results and analysis of recall and precision show that the manual seeds are of higher quality than the GDPR seeds, indicating that the seeding phase of pArborist significantly contributes to the derivation of privacy-revealing queries. Please note that our datasets are labeled in a privacy-conscious way which leads to approximately $5\%$ lower estimates of recall and precision than their actual values. Therefore, given (1) the over $90\%$ recall and the over $93\%$ precision in the default configuration in the Check-in and APP experiments and (2) the significant advantage of pArborist compared to FQID w.r.t. recall and precision in the Election experiment, we conclude that the performance of pAborist is satisfactory.

\subsubsection{Efficiency}
Figure \ref{figure_elec} illustrates a staircase-wise behavior of the used CPU time. The results indicate that deriving privacy-revealing queries is more efficient with the Check-in dataset than with the APP dataset. For Check-in, the CPU time used for deriving one privacy-revealing query in database scenarios is approximately $232.1$ ms with sufficient data and $371.4$ ms with insufficient data. Recall in database scenarios reaches its maximum of $93\%$ after $9.75$ seconds with sufficient data and $15.6$ seconds with insufficient data. In the data stream scenario, after a warm-up that takes approximately $900$ ms, each new detection takes approximately $1$ ms. The recall reaches $94\%$ after $0.92$ seconds. In the APP dataset, for database scenarios, it takes approximately $275.4$ ms with sufficient data and $393.5$ ms with insufficient data to derive a new privacy-revealing query. The maximum recall reaches $94\%$ after $12.7$ seconds and $18.1$ seconds for sufficient and insufficient data. The warm-up for data stream scenario takes $940$ ms, and each new derivation takes $1.6$ ms. The recall reaches $93\%$ after $0.94$ seconds.

Analyzing the results, we notice that the staircase-wise behavior of the used CPU time is led by the difference between Tree Growth and Node Bypassing steps. This confirms the different computational complexities of pArborist in these two steps. More CPU time is used for the experiments on the APP dataset, because the data tuples and seed queries for these experiments are more complex than those for the Check-in dataset. GDPR formulates simpler seed queries than manual seeds because the GDPR regulations usually correspond to simpler complex events, and thus simpler queries with fewer predicates. Considering the experiments on the Check-in dataset, GDPR seeds usually aim to query general events and thus usually consist of two or three A-F pairs, i.e., two or three predicates. For example, a GDPR seed that queries locations consists of two A-F pairs regarding longitude and latitude. Meanwhile, if we manually generate seeds based on the Check-in dataset, their queried data are usually more specific, and thus more complex. For example, a typical seed that is manually generated on the Check-in dataset can consist of four A-F pairs, querying the events that Bob went to Hotel H on April 10th and gave a positive review.

In summary, we see that pArborist reaches $94\%$ recall within $20$ seconds for database scenarios. The computation time pArborist introduces to data stream processing is on average $1.3$ ms for deriving a new privacy-revealing query, after an average of $920$ ms warm-up. This confirms the feasibility of pArborist for database scenarios and real-time decision-making for data stream scenarios.

\subsubsection{Practicality}
We have demonstrated that pArborist is applicable to various practical scenarios, including real-time processing for data streams, and provides satisfactory performance on real-world datasets, thereby validating the practicality of pArborist. In addition, Figures \ref{figure_recall} and \ref{figure_precision} indicate that six manual seeds provided by data producers are sufficient for pArborist to detect more than $90\%$ privacy-revealing queries with precision over $93\%$. If six seeds are still not feasible for data producers with marginal privacy expertise, data producers can directly employ pArborist based on GDPR regulations without any requirements for privacy expertise or additional inputs. A preliminary empirical study \cite{mastename} was conducted to collect user experience on pArborist from 21 graduate-level informatics students and ten participants without informatics background. In general, most of the participants did not perceive pArborist as difficult to utilize. They found the privacy-revealing queries generated by pArborist to be highly insightful and reported that these queries significantly changed their perceptions of privacy in multiple aspects. For example, several participants noted that pArborist detected privacy-revealing complex events that they had previously overlooked. We conclude that, in typical use cases, pArborist is beneficial for data producers with little privacy expertise and provides satisfactory performance and efficiency.

\section{Conclusion}
This work simplifies the process to label privacy-revealing complex events in data sharing by our novel semi-automatic approach pArborist. Evaluation with three real-world datasets demonstrates that pArborist achieves (1) high recall (> 90\%) and high precision (> 93\%), and (2) state-of-the-art performance compared to FQID \cite{podlesny2023quasi}. This work is intended to have the following impacts: (1) Several advanced privacy-enhancing technologies (PETs) require labeled privacy-revealing complex events as input to focus their obfuscation efforts onto these complex events. They achieve a superior privacy-utility trade-off compared to traditional methods. The introduction of pArborist empowers any data publisher to define these complex events and apply sophisticated PETs effectively. (2) With pArborist, data producers can proactively prevent unforeseen privacy breaches, avert legal liabilities, and mitigate economic risks, incentivizing data sharing for research and development purposes. (3) Due to the absence of publicly available datasets featuring labeled privacy-revealing complex events, we undertook the task of manually labeling the data for evaluation. We make these labeled data available to support other researchers and facilitate the development of benchmarking suites \cite{parborist}.

Our ongoing and future work addresses two challenges: (1) We are conducting thorough empirical studies on participants with varying levels of privacy expertise, aiming to augment the quantitative analysis of practicality in this paper with qualitative insights. (2) We seek to address data distribution drift in data streams by leveraging their underlying model and integrating reinforcement learning techniques. These efforts can enhance the robustness and adaptability of pArborist across various operational environments. 

\begin{acks}
This work was funded by the Parrot Project (Research Council of Norway, project number 311197).
\end{acks}

\bibliographystyle{ACM-Reference-Format}
\bibliography{sample-base}

@String{Computing = "Computing" }

@String{Computer = "{IEEE} Computer" }

@String{Springer = "Springer-Verlag" }

@phdthesis{podlesny2023quasi,
  title={Quasi-identifier discovery to prevent privacy violating inferences in large high dimensional datasets},
  author={Podlesny, Nikolai Jannik},
  year={2023},
  school={Universit{\"a}t Potsdam}
}

@inproceedings{hua2021method,
  title={A Method for Finding Quasi-identifier of Single Structured Relational Data},
  author={Hua, Yi and Li, Zhangbing and Sheng, Hankang and Wang, Baichuan},
  booktitle={2021 7th IEEE Intl Conference on Big Data Security on Cloud (BigDataSecurity), IEEE Intl Conference on High Performance and Smart Computing,(HPSC) and IEEE Intl Conference on Intelligent Data and Security (IDS)},
  pages={93--98},
  year={2021},
  organization={IEEE}
}

@inproceedings{palanisamy2018preserving,
  title={Preserving privacy and quality of service in complex event processing through event reordering},
  author={Palanisamy, Saravana Murthy and D{\"u}rr, Frank and Tariq, Muhammad Adnan and Rothermel, Kurt},
  booktitle={Proceedings of the 12th ACM International Conference on Distributed and Event-based Systems},
  pages={40--51},
  year={2018}
}

@inproceedings{gu2023differential,
  title={Differential Privacy for Protecting Private Patterns in Data Streams},
  author={Gu, He and Plagemann, Thomas and Benndorf, Maik and Goebel, Vera and Koldehofe, Boris},
  booktitle={2023 IEEE 39th International Conference on Data Engineering Workshops (ICDEW)},
  pages={118--124},
  year={2023},
  organization={IEEE}
}

@article{carvalho2022survey,
  title={Survey on privacy-preserving techniques for data publishing},
  author={Carvalho, T{\^a}nia and Moniz, Nuno and Faria, Pedro and Antunes, Lu{\'\i}s},
  journal={arXiv preprint arXiv:2201.08120},
  year={2022}
}

@article{shu2015privacy,
  title={Privacy-preserving detection of sensitive data exposure},
  author={Shu, Xiaokui and Yao, Danfeng and Bertino, Elisa},
  journal={IEEE transactions on information forensics and security},
  volume={10},
  number={5},
  pages={1092--1103},
  year={2015},
  publisher={IEEE}
}

@inproceedings{nan2015uipicker,
  title={UIPicker:User-Input privacy identification in mobile applications},
  author={Nan, Yuhong and Yang, Min and Yang, Zhemin and Zhou, Shunfan and Gu, Guofei and Wang, XiaoFeng},
  booktitle={24th USENIX Security Symposium (USENIX Security 15)},
  pages={993--1008},
  year={2015}
}

@inproceedings{caliskan2014privacy,
  title={Privacy detective: Detecting private information and collective privacy behavior in a large social network},
  author={Caliskan Islam, Aylin and Walsh, Jonathan and Greenstadt, Rachel},
  booktitle={Proceedings of the 13th Workshop on Privacy in the Electronic Society},
  pages={35--46},
  year={2014}
}

@inproceedings{tesfay2019privacybot,
  title={Privacybot: detecting privacy sensitive information in unstructured texts},
  author={Tesfay, Welderufael B and Serna, Jetzabel and Rannenberg, Kai},
  booktitle={2019 sixth international conference on social networks analysis, management and security (SNAMS)},
  pages={53--60},
  year={2019},
  organization={IEEE}
}

@inproceedings{mehdy2019privacy,
  title={Privacy disclosures detection in natural-language text through linguistically-motivated artificial neural networks},
  author={Mehdy, Nuhil and Kennington, Casey and Mehrpouyan, Hoda},
  booktitle={Security and Privacy in New Computing Environments: Second EAI International Conference, SPNCE 2019, Tianjin, China, April 13--14, 2019, Proceedings 2},
  pages={152--177},
  year={2019},
  organization={Springer}
}

@inproceedings{kopeykina2019automatic,
  title={Automatic privacy detection in scanned document images based on deep neural networks},
  author={Kopeykina, Lyudmila and Savchenko, Andrey V},
  booktitle={2019 International Russian Automation Conference (RusAutoCon)},
  pages={1--6},
  year={2019},
  organization={IEEE}
}

@article{vasalou2011privacy,
  title={Privacy dictionary: A new resource for the automated content analysis of privacy},
  author={Vasalou, Asimina and Gill, Alastair J and Mazanderani, Fadhila and Papoutsi, Chrysanthi and Joinson, Adam},
  journal={Journal of the American Society for Information Science and Technology},
  volume={62},
  number={11},
  pages={2095--2105},
  year={2011},
  publisher={Wiley Online Library}
}

@article{du2009confidence,
  title={Confidence interval or p-value?: part 4 of a series on evaluation of scientific publications},
  author={Du Prel, Jean-Baptist and Hommel, Gerhard and R{\"o}hrig, Bernd and Blettner, Maria},
  journal={Deutsches {\"A}rzteblatt International},
  volume={106},
  number={19},
  pages={335},
  year={2009},
  publisher={Deutscher Arzte-Verlag GmbH}
}

@inproceedings{yang2019revisiting,
  title={Revisiting user mobility and social relationships in LBSNs: a hypergraph embedding approach},
  author={Yang, Dingqi and Qu, Bingqing and Yang, Jie and Cudre-Mauroux, Philippe},
  booktitle={The world wide web conference},
  pages={2147--2157},
  year={2019}
}

@article{yang2020lbsn2vec++,
  title={LBSN2VEC++: Heterogeneous hypergraph embedding for location-based social networks},
  author={Yang, Dingqi and Qu, Bingqing and Yang, Jie and Cudr{\'e}-Mauroux, Philippe},
  journal={IEEE Transactions on Knowledge and Data Engineering},
  volume={34},
  number={4},
  pages={1843--1855},
  year={2020},
  publisher={IEEE}
}

@article{cohen1960coefficient,
  title={A coefficient of agreement for nominal scales},
  author={Cohen, Jacob},
  journal={Educational and psychological measurement},
  volume={20},
  number={1},
  pages={37--46},
  year={1960},
  publisher={Sage Publications Sage CA: Thousand Oaks, CA}
}

@article{kendall1938new,
  title={A new measure of rank correlation},
  author={Kendall, Maurice G},
  journal={Biometrika},
  volume={30},
  number={1-2},
  pages={81--93},
  year={1938},
  publisher={Oxford University Press}
}

@article{koch2004intraclass,
  title={Intraclass correlation coefficient},
  author={Koch, Gary G},
  journal={Encyclopedia of statistical sciences},
  year={2004},
  publisher={Wiley Online Library}
}

@article{articleocc,
author = {Kim, Albert and Escobedo-Land, Adriana},
year = {2015},
month = {07},
pages = {},
title = {OkCupid Data for Introductory Statistics and Data Science Courses},
volume = {23},
journal = {Journal of Statistics Education},
doi = {10.1080/10691898.2015.11889737}
}

@article{sabuncu10usa,
  title={USA Nov. 2020 Election 20 Mil. Tweets (with Sentiment and Party Name Labels) Dataset (2020)},
  author={Sabuncu, IBRAHIM},
  year={2020},
  journal={URL https://dx. doi. org/10.21227/25te-j338}
}

@book{luckham2002power,
  title={The power of events},
  author={Luckham, David},
  volume={204},
  year={2002},
  publisher={Addison-Wesley Reading}
}

@misc{regulation2018general,
  title={General Data Protection Regulation (GDPR)--Legal Text},
  author={Regulation, General Data Protection},
  year={2018},
  publisher={General Data Protection Regulation (GDPR). https://gdpr-info. eu}
}

@software{parborist,
  author = {He Gu},
  doi = {},
  month = {12},
  title = {{pArborist}},
  url = {https://github.com/eivgreen/pArborist},
  version = {1.0.0},
  year = {2024}
}

@article{lotfian2025pattern,
  title={Pattern-Level Privacy Protection in Event-Based Systems},
  author={Lotfian Delouee, Majid and Degeler, Victoria and Amthor, Peter and Schut, Martijn C and Koldehofe, Boris},
  journal={SN Computer Science},
  volume={6},
  number={8},
  pages={1014},
  year={2025},
  publisher={Springer}
}

@mastersthesis{mastename,
  author  = "Sofie Mjaaland",
  title   = "Data Patterns and Privacy: How Non-Expert Users Make Sense of the pArborist Algorithm",
  school  = "University of Oslo",
  year    = "2026",
  type    = "Master Thesis",
}


\end{document}